\documentclass{elsarticle}

\usepackage{hyperref,subfigure}

\journal{Information Sciences} 

\usepackage{xcolor}
\usepackage[normalem]{ulem}  
\usepackage[american]{babel}
\usepackage{comment} 
\usepackage{amssymb}
\usepackage{amsmath}
\usepackage[utf8]{inputenc}
\usepackage{booktabs}
\usepackage{cases}
\usepackage{xspace}
\usepackage{enumerate}
\usepackage{graphicx}
\usepackage[bold]{hhtensor}
\usepackage{rotating}

\newcommand{\gijon}{{Gijón}\xspace}
\newcommand{\barcelona}{{Barcelona}\xspace}
\newcommand{\madrid}{{Madrid}\xspace}
\newcommand{\nyc}{{New York}\xspace}
\newcommand{\paris}{{Paris}\xspace}
\newcommand{\london}{{London}\xspace}

\newcommand{\oursystem}{\emph{ELVis}\xspace}
\newcommand{\oursystemlong}{\emph{Explaining Likings Visually}\xspace}

\newcommand{\argmax}[1]{\underset{#1}{\operatorname{argmax}\ }}

\newcommand{\usersset}{\ensuremath{\mathcal{U}}\xspace}
\newcommand{\itemsset}{\ensuremath{\mathcal{I}}\xspace}
\newcommand{\newop}[2]
	{\newcommand{#1}{\ensuremath{\operatorname{#2}}\xspace}}

\newop{\rs}{R}
\newop{\novelty}{novelty}
\newop{\popularity}{popularity}
\newop{\weight}{weight}
\newop{\reg}{reg}
\newop{\users}{users}
\newop{\items}{items}
\newop{\photos}{photos}

\begin{document}

\begin{frontmatter}

\title{Towards Explainable Personalized Recommendations by Learning from Users' Photos}

\author{Jorge Díez, Pablo Pérez-Núñez, Oscar Luaces, Beatriz Remeseiro, Antonio Bahamonde}
\address{Artificial Intelligence Center \\ Universidad de Oviedo at Gijón \\ Campus de Viesques, 33204, Gijón, Spain}




\begin{abstract}
Explaining the output of a complex system, such as a Recommender System (RS), is becoming of utmost importance for both users and companies. In this paper we explore the idea that personalized explanations can be learned as recommendation themselves. There are plenty of online services where users can upload some photos, in addition to rating items. We assume that users take these photos to reinforce or justify  their opinions about the items. For this reason we try to predict what photo a user would take of an item, because that image is the argument that can best convince her of the qualities of the item. In this sense, an RS can explain its results and, therefore, increase its reliability. Furthermore, once we have a model to predict attractive images for users, we can estimate their distribution. Thus, the companies acquire a vivid knowledge about the aspects that the clients highlight of their products. The paper includes a formal framework that estimates the authorship probability for a given pair (user, photo). To illustrate the proposal, we use data gathered from TripAdvisor containing the reviews (with photos) of restaurants in six cities of different sizes.
\end{abstract}


\end{frontmatter}

\section{Introduction}\label{sec:introduction}




Explainable Artificial Intelligence (XAI) is becoming an important area of interest since explainability is increasingly necessary to meet stakeholder demands. In particular, the General Data Protection Regulation (GDPR) \cite{voigt2017gdpr} of the European Union demands transparency in systems that take decisions affecting people, making explanations more needed than ever. Additionally, explanations may help increase the trust of users in AI algorithms, since people rely not only on their efficacy but also on the degree of understanding of the process they follow.

{Recommender Systems (RS) are methods that suggest items to users \cite{ricci2011introduction}, based on item popularity (general recommendations) or on users' tastes (personalized recommendations). Since they provide suggestions to users, explainability plays an important role on them. There are different ways to help users assimilate the outputs of complex systems such as RS. Some attempts of explainable recommendations can be found in the literature, including the use of visualizing trees of items \cite{hernando2013trees} or the use of item tags \cite{zheng2019explore}.}

The central point of our research relies on the use of a collaborative platform in which there are items and users, where users can express their opinions about items. These opinions, in addition to ratings, include photos taken by users and uploaded to the platform. We want to exploit the popular saying that a picture is worth a thousand words. In this sense, our idea is that a recommendation of an item can be accompanied by an image that explains it, showing the item's most relevant characteristics for the user; such image should be the most adequate to justify the user's ratings.



For this purpose, we present a method to learn personalized explanations based on visual information, called \oursystem (\oursystemlong). Our approach is not aimed at predicting whether a user would like or not an item; that could be done by a \emph{conventional} RS.
Instead, we aim to recommend photos in order to provide a personalized cover picture of an item; thus, making it more attractive for users. 

From a formal point of view, we will predict a photo that a user would upload. The justification for this approach is that  with their photos, users want to highlight the most relevant aspects of the items according to their tastes. They take the photos to explain to everyone how accurate and balanced their ratings are; thus, their own photos are the reason that convinces them the most.

There is another {aspect} that we want to remark here, which is the usefulness of this proposal seen from the side of companies. Once we have a model to predict images, we can estimate the distribution of those predicted images. This {distribution} can {provide very useful information} about the aspects of products that {most attract} the attention of the customers.

Finally, we would like to emphasize that the main contributions of our research are three: 1) the conceptual framework according to which the personalized explanations are reasons that must be learned, such as the recommended items themselves, 2) the datasets\footnote{The datasets and the code will be publicly available with the publication of the paper} used in the experiments, and 3) the selection of the most attractive images of an item (or a set of them) for the users as a whole, what explains the features that customers highlight most of an item.

The remainder of this manuscript is structured as follows. Section \ref{sec:related_work} includes other proposals that deal with images and/or use data from popular online platforms. Section~\ref{sec:framework} is devoted to introduce the formal framework to suggest the photo that the user would take of an item, instead of the rating that the user would assign to it. In that section we also underscore that both learning tasks are conceptually the same, and so both of them can be tackled as a binary classification task. Section \ref{sec:topology} presents the network used to solve our binary classification problem by learning on pairs of users and photos. Section \ref{sec:results} describes some experiments carried out to show the advantages of our approach compared with a couple of baselines. In these experiments we used public TripAdvisor data on restaurants in six cities that range from medium to very large sizes. Finally, Section \ref{sec:conclusions} closes the manuscript with our conclusions and future lines or research.

\section{Related work}
\label{sec:related_work}

This section includes a brief review of some approaches that use visual information to provide recommendations. On the one hand, we highlight some methods for video or movie recommendations that select the most appealing thumbnail or cover, respectively, some of which are based on popular platforms such as YouTube or Netflix. On the other hand, we stand out some approaches for restaurant recommendations, with an special mention to another popular platform such as TripAdvisor. Finally, we present some state-of-art methods that use photos to provide recommendations, and the rationale of our approach compared to them.

\subsection{YouTube}

In the context of videos, YouTube is the largest RS in the industrial world. YouTube recommendations are described in \cite{covington2016deep}, including details about its design and maintenance. The complete system is a deep learning approach composed by two neural networks: one to generate a list of video candidates based on the users' activity history, and the other  one to rank them in a personalized way. Note that in the second stage the authors proposed a deep collaborative filtering model where both videos and users are represented by means of rich feature sets of descriptors, instead of using matrix factorization \cite{koren2009matrix}, as in previous approaches.

Regarding video retrieval, Liu et al. \cite{liu2015multi} proposed a multi-task approach to automatically select query-dependent video thumbnails. They are selected from video frames, based on visual and side information. The authors extract visual information by means of a CNN architecture, and side information from the query by a word embedding model (\emph{GloVe} \cite{pennington2014glove}). Next, the two vector representations are mapped into a latent semantic space, in which the relevance of thumbnails can be estimated for the final selection.

\subsection{Netflix}

Netflix platform is endowed with a set of recommender tools. An overview of the algorithms used together with their business value is presented in \cite{gomez2016netflix}. Among them, we would like to highlight how Netflix provides personalized covers of the available contents \cite{netflix16selecting}. The idea lies in exploiting movies' and users' information to select the picture that best represents a movie across all users.

Given the high diversity in users' preferences, Amat et al. \cite{amat2018artwork} proposed to personalize the best picture per movie for each individual user. Their target was how to convince users to watch a movie, by showing them some visual evidence that supports the recommendation. The approach is based on contextual bandits and online machine learning, and it selects the images and makes the personalized recommendations as part of the same process.

Other works can be found in the literature applied to movie recommendations, which were evaluated on widely used datasets such as MovieLens and Netflix Prize. For example, Ortega et al. \cite{ortega2016recommending} proposed a recommender system based on matrix factorization and collaborative filter to provide movie re\-commendations to groups of users.

\subsection{TripAdvisor}

TripAdvisor is a very popular (not personalized) recommendation platform of the hospitality sector, in which users upload their opinions about restaurants, including ratings, text reviews and photos. In this context, deep learning networks have been used to improve users' experience by showing them the most appealing pictures, but not in a personalized manner \cite{amis17improving}. The authors gathered the training data from the platform, and then manually selected thousands of \emph{preference judgments}; that is, pairs to learn a ranking of photos for given a property. Regarding the model architecture, they used a \emph{siamese network} built on the top of the ResNet50 \cite{he2016deep}, thus allowing to learn on pairs of photos.

Chu and Tsai \cite{chu2017hybrid} designed a study in which 
restaurant attributes and users preferences are both represented by visual features, allowing to link content-based and collaborative filtering.
The process with images uses standard embedding CNN procedures with additional ad hoc features. In order to deal with several photos, the authors use averaging or maximum aggregations. 

Focusing on restaurant recommenders that do not use images, several works found in the literature are evaluated with TripAdvisor data. Some recent examples are a decision support model \cite{zhang2017novel} that recommends restaurants to tourists making use of social information, including online reviews and social relationships; and a recommender system \cite{zhang2018personalized} that considers group correlations for both users and restaurants.

Finally, it is worth noting that there are also hotel recommenders that use TripAdvisor. For example, Nilash et al. \cite{nilashi2018travelers} presented a fuzzy-based approach that uses multi-criteria ratings extracted from online reviews to provide hotel recommendations. Another example is the one proposed by Liu et al. \cite{liu2019time} that recommends hotels taking into account the time in which the reviews were made by users and, thus, possible changes in their behaviors and tastes.

\subsection{Photos and recommendations}

To our knowledge, the first attempt to use photos to provide recommendations is the content-based RS presented in  \cite{he2016vbpr}, called VBPR (Visual Bayesian Personalized Ranking). It is a factorization system in which the description of the restaurants (items) are the features learned by a CNN from one single image of the item.


Kang et al. \cite{kang2017visually} seek to extend the previous contribution in \cite{he2016vbpr}, showing that recommendation performance can be significantly improved by learning \emph{fashion aware} image representations directly, i.e., by training the image representation (from the pixel level) and the RS jointly. 
In a more recent work, Tang et al. \cite{tang2019adversarial} focused their attention on the lack of robustness that affects to VBPR. In this sense, the authors proposed an approach based on adversarial learning to obtain a robust multimedia recommender, which was evaluated on Pinterest and Amazon datasets.

\subsection{Rationale of the approach}

Other recent works have explored the use of images in the context of RS; however, to the best of our knowledge, there are no attempts, in the literature or commercially, able to predict the photos that the user would take with the main aim of providing explainable recommendations. Our target is to provide a personalized explanation drawn from the core of the RS that learns the interest of users in items. For this purpose, users' photos are employed. 

Regarding other systems that recommend images, such as the one previously described \cite{tang2019adversarial}, there are two main differences that deserve to be highlighted. First, images handled in our proposal are photos uploaded by the users of a platform to show their opinion about an item, instead of being uploaded by its responsible in commercial terms; thus, we have to face a challenge in terms of the quality of the images and the variety of objects and information displayed in them. Secondly, photos in our case can be organized by the user that uploaded them or by the item they correspond
; thus, obtaining a complex structure of images that we have to handle properly. Finally, our objective is to give an explanation, which makes us explore the set of images in a peculiar way.

With respect to restaurant recommenders, users that look for restaurants in popular platforms are interested, not only in the rating given by other users and their comments, but also in their photos, taken mainly of food, restaurant atmosphere and restaurant location. For this reason, some RS found in the literature, such as the work presented in \cite{chu2017hybrid} previously mentioned, use this kind of information. However, all of them use photos to compute a personalized list of recommendations, while our target is to predict photos as a way to provide personalized explanations.

\section{General framework}\label{sec:framework}

%
%
%
%
%

Let us consider a set of users \usersset and a set of items \itemsset. Typically in a collaborative recommender platform, we have some pairs with labels

\begin{equation}\label{eq:rec_likes}
(\vec{u}, \vec{it}) \leadsto 0|1,
\end{equation}
where $\vec{u} \in \usersset$ and $\vec{it} \in \itemsset$. 
The label 0 means that the user does not like the item, while 1 stands for the opposite; that is, these labels represent the \emph{reaction} of the user $\vec{u}$ about the item $\vec{it}$.

Nonetheless, in this paper we  consider a set of photos taken by users to emphasize some aspects of the items. Formally, we have a collection of sets:
\begin{itemize}
    \item $\text{photos}(\vec{u})$: photos taken by user $\vec{u} \in \usersset$,
    \item $\text{photos}(\vec{it})$: photos of item $\vec{it} \in \itemsset$, and
    \item $\text{photos}(\vec{u},\vec{it})$: $\text{photos}(\vec{u}) \cap \text{photos}(\vec{it})$.
\end{itemize}
Our interest is to be able to detect the photos taken by a given user since we understand that they are representative of that user's tastes. Therefore, we assume to have a set of labeled pairs

\begin{equation}
    \label{eq:rec_photos}
    (\vec{u}, \vec{f}) \leadsto
    \left\{
    \begin{array}{@{}
    ll}
        0, & \vec{f}\notin\photos(\vec{u}) \\
        1, & \vec{f}\in\photos(\vec{u})
    \end{array}{}
    \right.
\end{equation}
where $\vec{u}\in\usersset$ denotes a user, $\vec{f}\in\photos(\vec{it})$ is a photo of an item, and the label point out the \emph{authorship} of the photos. Thus, the label will be 0 to indicate that the photo was not taken by user $\vec{u}$ and 1 otherwise.

As usual in RS, we are going to estimate the probabilities for positive labels. The main difference is that, in this paper, labels' examples represent the authorship of the photos. We assume that a user takes photos of an item trying to reflect the most important characteristics that make the user like/dislike the item. Thus, we want to learn a procedure to select the photo of an item $\vec{it}$ that best represents the tastes of a user $\vec{u}$, that is,

\begin{equation}
    \label{eq:f*}
    \vec{f^*} = \argmax{\vec{f} \in \text{photos}(\vec{it})}{\Pr(\vec{u}, \vec{f})}.
\end{equation}
Therefore, we aim at solving a binary classification task to estimate $\Pr(\vec{u}, \vec{f})$. The rationale is that once we learn to predict the probability of a photo to be taken by a user $\vec{u}$, we can apply that model to select photos taken by other users, but with a high probability to be also taken by $\vec{u}$. This will allow us to show to $\vec{u}$ the most adequate photos of new items, those that the user would have probably taken.

This idea can be generalized to a group of users by defining a compatibility function of a photo, $\vec{f}$, with a set of users $\mathcal{S} \subseteq \usersset$ according to their tastes by

\begin{equation}
    \label{eq:SU_photo}
    \phi(\mathcal{S},\vec{f}) = \sum_{\vec{u} \in \mathcal{S}} \Pr(\vec{u},\vec{f}).
\end{equation}

Therefore, the photo $\vec{f^*}$ that best \emph{explains the tastes of a group $\mathcal{S}$ of users} regarding a given item $\vec{it}$ can be obtained by

\begin{equation}\label{eq:SU_SI_photo}
\vec{f}^* = \argmax{\vec{f}\in \text{photos}(\vec{it})} \phi(\mathcal{S},\vec{f}).
\end{equation}

\subsection{Dataset definition} \label{sec:formal_dataset}

The definition of training and test sets is extremely important in order to focus the learning task towards our objective, which is to learn the probability of a photo to be taken by a user. The following steps describe the sequence to obtain the datasets of photos and users (graphically depicted in Figure~\ref{fig:train_test_split}) starting from the raw data, and can be considered as part of the learning task previously described in this section:

\begin{figure}[htbp]
    \centering
    \includegraphics[width=\linewidth]{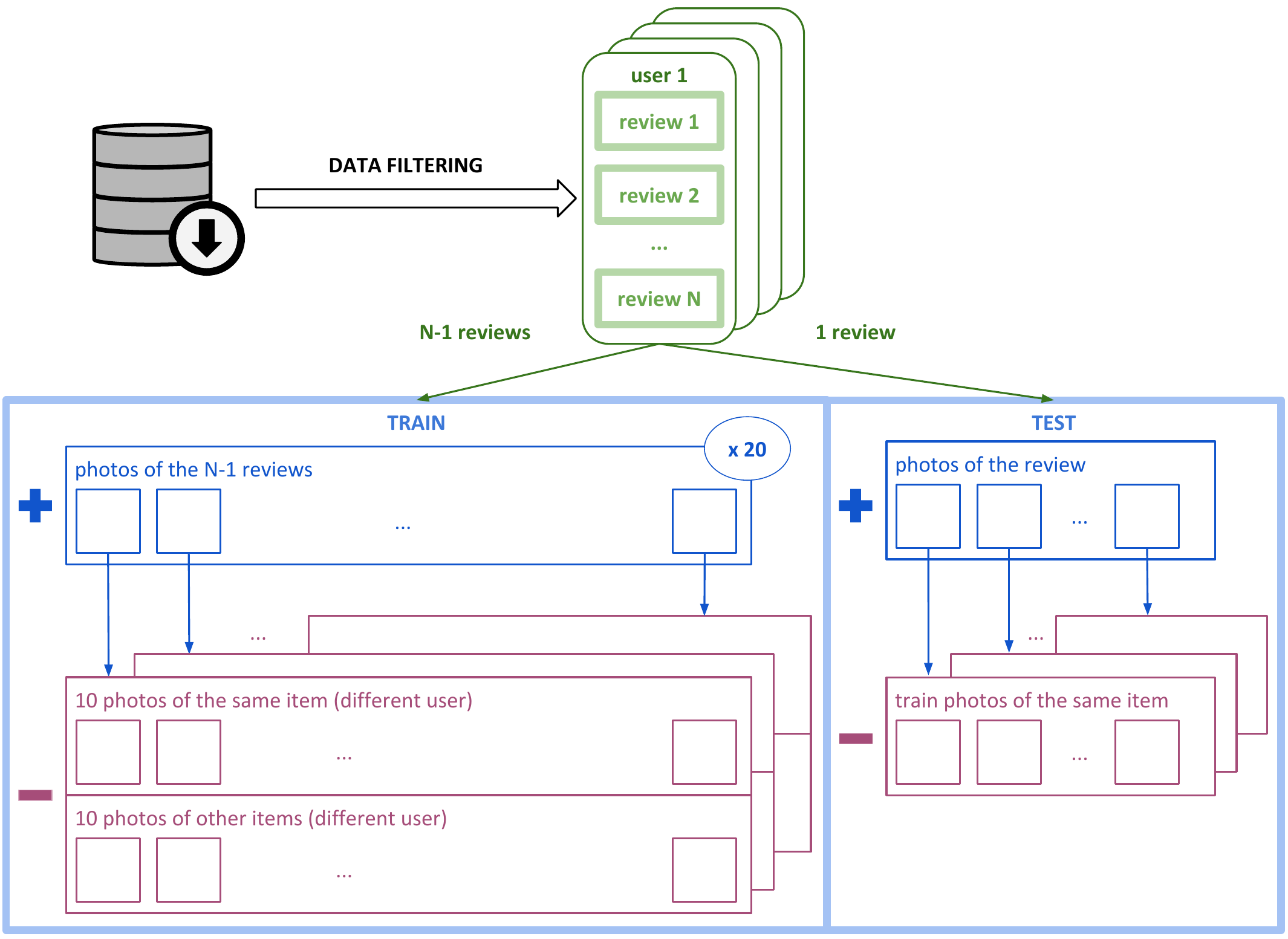}
    \caption{Reviews of all items are downloaded, and next filtered to get only those reviews with users' photos. For the sake of simplicity, if any user has more than one review for the same item, only the most updated one is maintained. As a result, we have a set of users, each one with $N$ reviews corresponding to $N$ different items. In order to split these data into training and test sets, $1$ review is used for testing and the rest ($N-1$) are used for training. Training set: the positive samples correspond to all the photos of the $N-1$ reviews, each one repeated 20 times (oversampling); while the negative samples are added by selecting photos taken by other users, 10 from the same item and 10 from other items. Test set: the positive samples correspond to all the photos from the review, while the negative samples are added by selecting all the photos of the same item in the training set.}
    \label{fig:train_test_split}
\end{figure}

\begin{enumerate}
    \item Data gathering: The first step is to gather the data. Usually, these data are items' reviews (with or without photos)  provided by the users of a given service.
    \item Data filtering: Only reviews with photos are considered for the dataset. Additionally, if any user has two or more reviews for the same item, only the most recent one is considered. Thus, we obtain a set of users, each one with $N$ reviews corresponding to $N$ different items. Notice that not all users have the same number $N$ of reviews: most of the users rate only one or a few items, while only a few users rate a lot of articles. 
    \item Train/Test split: For those users that have at least two reviews, one of them  is used for testing purposes, and the remaining $N-1$  are reserved for training. In this manner, the positive samples of both training and test sets are composed of pairs of users and photos $(\vec{u}, \vec{f})$ with positive labels.
    \item Negative samples in the training set: For each pair $(\vec{u}, \vec{f})$, which represents that user $\vec{u}$ took the photo $\vec{f}$ of an item $\vec{it}$, we add up 10 negative pairs $(\vec{u}, \vec{f'})$, where  $\vec{f'}$ is a photo of the same item $\vec{it}$  but not taken by  user $\vec{u}$; and 10 negative pairs $(\vec{u}, \vec{f"})$, where $\vec{f"}$ is a photo of a different item (other than $\vec{it}$) also taken by a different user.
    \item Oversampling: In order to get a balanced training set, each positive sample $(\vec{u}, \vec{f})$ is repeated 20 times to compensate the 20 negative pairs added in previous step.
    \item Negative samples in the test set: For each pair $(\vec{u}, \vec{f})$, we add up negative pairs $(\vec{u}, \vec{f'})$ with all the photos $\vec{f'}$ of the same item in the training set.
\end{enumerate}

\section{Topology of the  network}
\label{sec:topology}

\begin{figure}[htbp]
    \centering
    \includegraphics[width=0.5\linewidth]{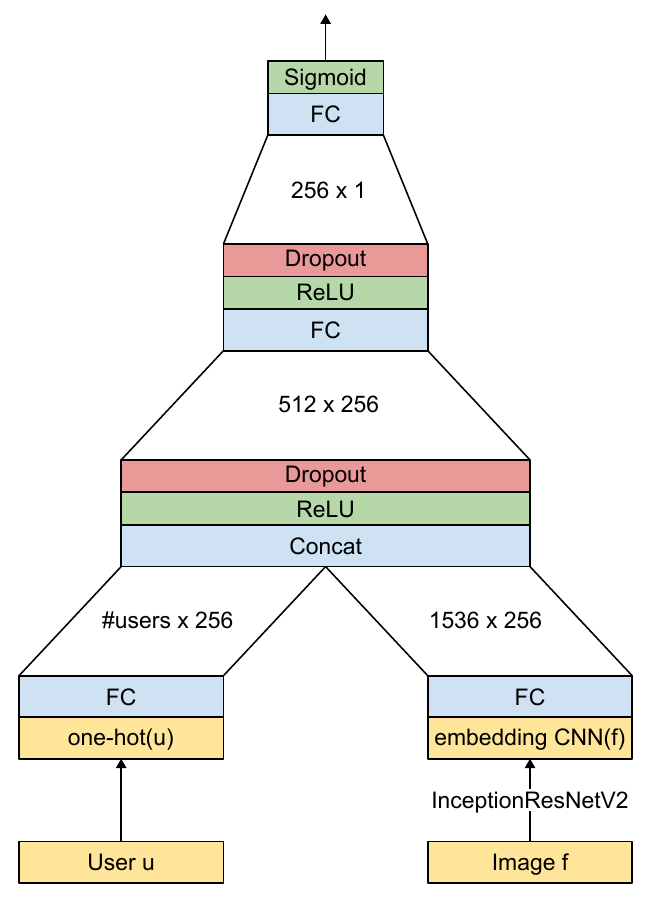}
    \caption{Topology of the network employed in \oursystem to learn $\Pr(\vec{u}, \vec{f})$ from labeled pairs of users and photos $(\vec{u}, \vec{f})$}
    \label{fig:topology}
\end{figure}

In this section, we present the network employed to induce the distribution of $\Pr(\vec{u}, \vec{f})$. Taking into account that we have to solve a binary classification task, the optimization of the \emph{binary cross-entropy} loss function was carried out using a model that learns on pairs of users and photos $(\vec{u}, \vec{f})$. Figure~\ref{fig:topology} depicts the architecture of the proposed model \oursystem.




The input data of \oursystem are codified as follows:
\begin{itemize}
    \item User $\vec{u}$. Users are represented by a \emph{one-hot} codification, and then mapped into a 256-dimensional embedding.

    \item Image $\vec{f}$. Photos are codified using a Convolutional Neural Network (CNN); more specifically, the convolutional base of the Inception-ResNet-v2 model \cite{szegedy2017inception} with weights pre-trained on ImageNet\footnote{https://keras.io/applications/\#inceptionresnetv2}. The embedding provided by the CNN, composed of 1,536 deep features, is next transformed by a Fully Connected (FC) layer to finally obtain a vector with 256 elements.
\end{itemize}

Once the input data are codified as previously described, the two vectors are concatenated and further processed by a sequence of different layers that include Fully Connected (FC), Rectified Linear Unit (ReLU) \cite{nair2010rectified}, and Dropout \cite{srivastava2014dropout}. The purpose of these layers is to learn a nonlinear function able to determine if a given photo was taken by a given user. The authorship is given in terms of the joint probability, $\Pr(\vec{u}, \vec{f})$, so finally a sigmoid activation function is used to produce a probability output in the range $[0,1]$.

We came up with this architecture after numerous empirical tests using other networks with variations in the number and size of the layers.

\section{Experimental results}\label{sec:results}

In this section we report the results obtained in some experiments carried out to test the performance of the algorithm \oursystem proposed in this paper. First, we present the datasets used. Then we describe the experiments in detail, including the two baseline methods used to compare with our method. Finally, we show and discuss the results obtained from different points of view.

\subsection{Datasets}
\label{sec:datasets}

We used real-world data collected in $2018$ and $2019$ from TripAdvisor reviews of restaurants in six cities around the world. We used three cities of Spain, including the two biggest cities in the country, \barcelona (population: 1.6 million) and \madrid (pop.: 3.2 million). The third one, \gijon, is a medium size city of around 300,000 inhabitants. In addition, we also used data from other big cities of the world, such as \nyc (8.3 million), \paris (2.1 million) and \london (8.9 million). 

In order to make a proper estimation of the performance of learning methods and a fair comparison among them, we split all the data crawled from TripAdvisor into training and test sets as explained in Section~\ref{sec:formal_dataset} and depicted in Figure~\ref{fig:train_test_split}. The basic statistics of the datasets obtained through this splitting procedure are shown in Table~\ref{tab:datasets}.






\begin{table}[tbp]
\begin{center}
    \begin{tabular}{l|rrr}
        \toprule
        & \multicolumn{3}{c}{All data}\\
        & \#Users & \#Rst & \#Photos \\
        \midrule
        \gijon       &  5,139 & 598 & 18,679 \\
        \barcelona   & 33,537 & 5,881 & 150,416 \\
        \madrid      & 43,628 & 6,810 & 203,905 \\
        \nyc         & 61,019 & 7,588 & 231,141 \\
        \paris       & 61,391 & 11,982 & 251,636 \\
        \london     & 134,816 & 13,888 & 479,798 \\
        \bottomrule
    \end{tabular}\\
    \hspace{5.5em}\includegraphics[height=2.5em,width=10em]{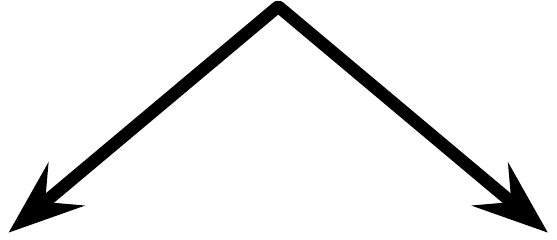}\\
    \begin{tabular}{l|rrr|rrr}
        \toprule
        & \multicolumn{3}{c}{Training Set} & \multicolumn{3}{|c}{Test Set}\\
        \midrule
        & \#Users & \#Rst & \#Photos & \#Users & \#Rst & \#Photos \\
        \gijon       & 5,139 & 598 & 16,302 & 1,023 & 346 & 2,377 \\
        \barcelona	& 33,537 & 5,881 & 130,674 & 8,697 & 3,211 & 19,742 \\
        \madrid  	& 43,628 & 6,810 & 176,763 & 11,874 & 3,643 & 27,142 \\
        \nyc	& 61,019 & 7,588 & 196,315 & 16,842 & 4,135 & 34,826 \\
        \paris   	& 61,391 & 11,982 & 219,588 & 15,242 & 6,345 & 32,048 \\
        \london	    & 134,816 & 13,888 & 416,356 & 30,393 & 8,097 & 63,442 \\
        \bottomrule
    \end{tabular}
    \caption{Basic statistics of the datasets used in our experiments. \emph{Rst} means restaurants in the column labels}
    \label{tab:datasets}
\end{center}
\end{table}

Figures~\ref{fig:datasets_fig1} and~\ref{fig:datasets_fig2} depict basic information of the whole data downloaded from the TripAdvisor site. The graphics display, for each city, the distribution of:
\begin{itemize}
    \item Restaurants regarding the number of photos taken on them.
    \item Restaurants regarding the number of reviews, which is equivalent to the number of users who visited them (and took photos).
    \item Reviews regarding their number of photos. Worth of mention is that our crawler just downloaded the four photos that TripAdvisor shows per review (in case there were some photos).
    \item Users regarding the number of restaurants they visited (reviews).
    \item Users regarding the number of photos taken.
\end{itemize}

\newcommand{\espacio}{\hspace{0.65ex}}
\newcommand{\cw}{0.32\columnwidth}

\begin{figure}[p]
    \centering
    \begin{tabular}{@{}c@{\espacio}c@{\espacio}c@{}}
        \gijon & \barcelona & \madrid \\
        \includegraphics[width=\cw, trim=25 5 40 40, clip]{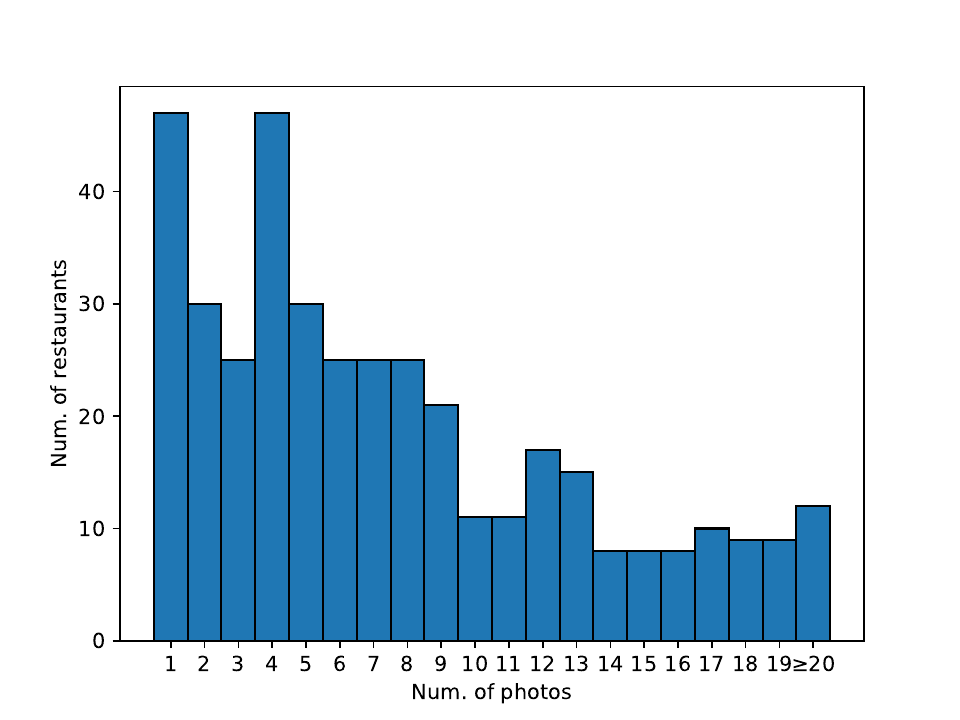}&
        \includegraphics[width=\cw, trim=15 5 40 40, clip]{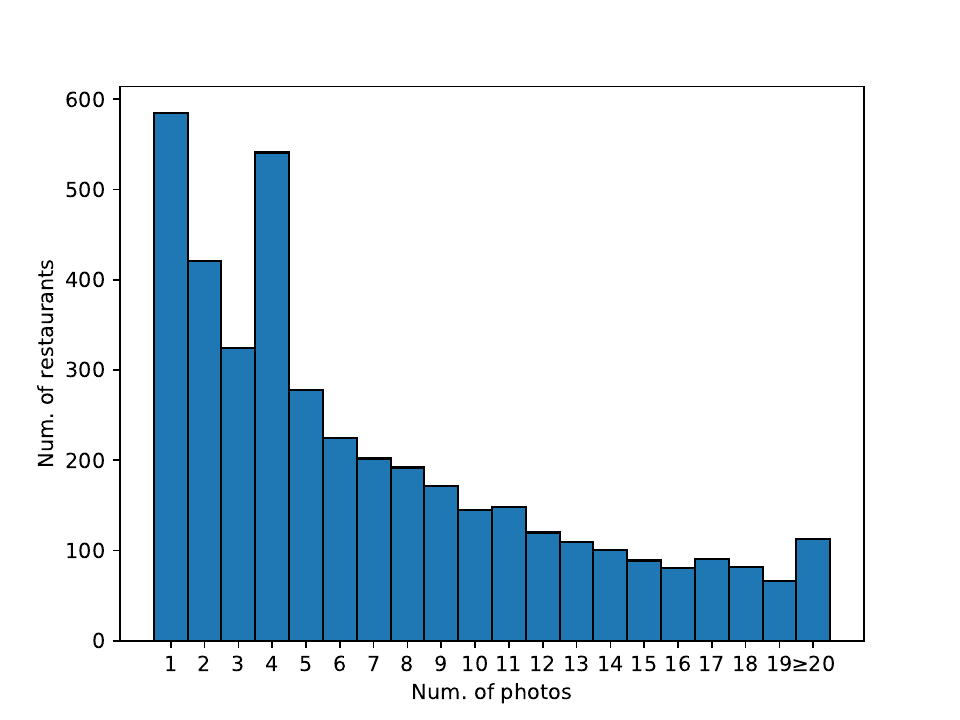}&
        \includegraphics[width=\cw, trim=15 5 40 40, clip]{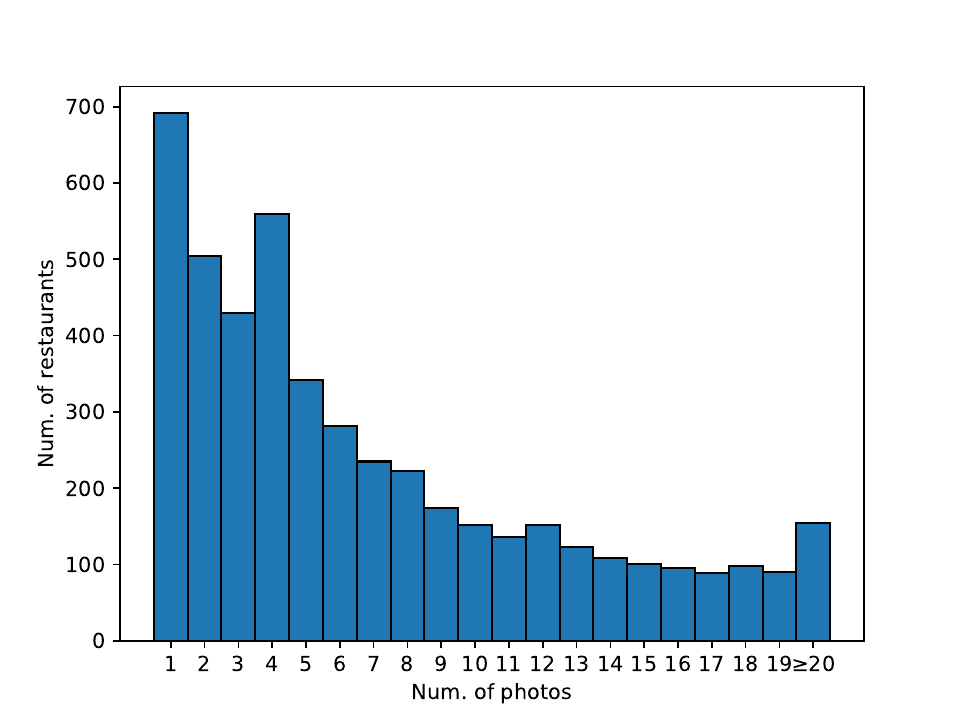}\\
        
        \includegraphics[width=\cw, trim=15 7 40 40, clip]{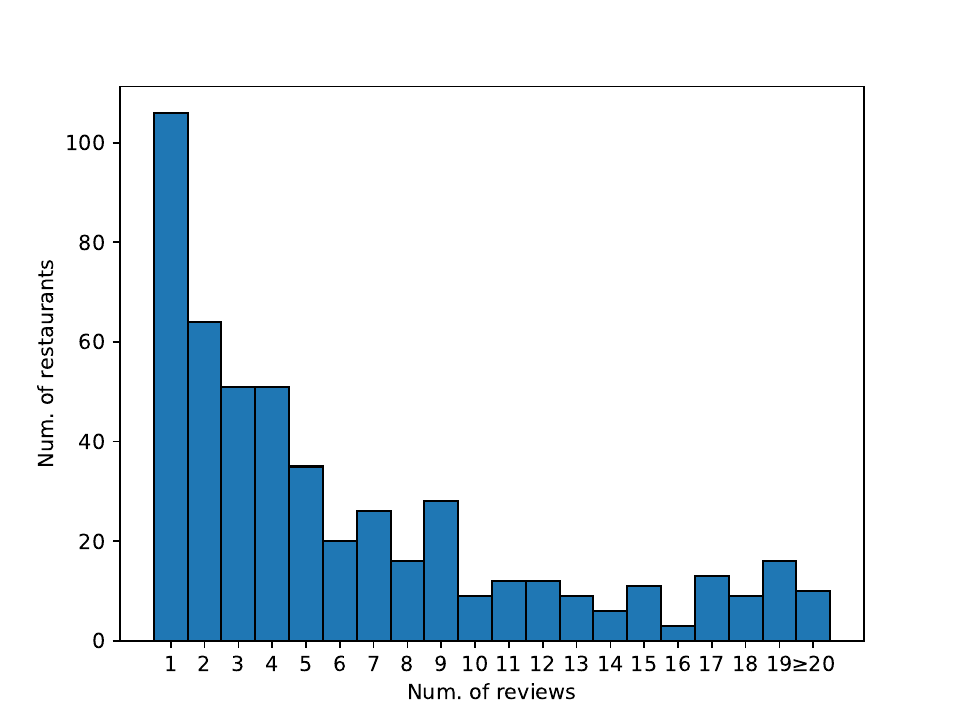}&
        \includegraphics[width=\cw, trim=10 5 40 40, clip]{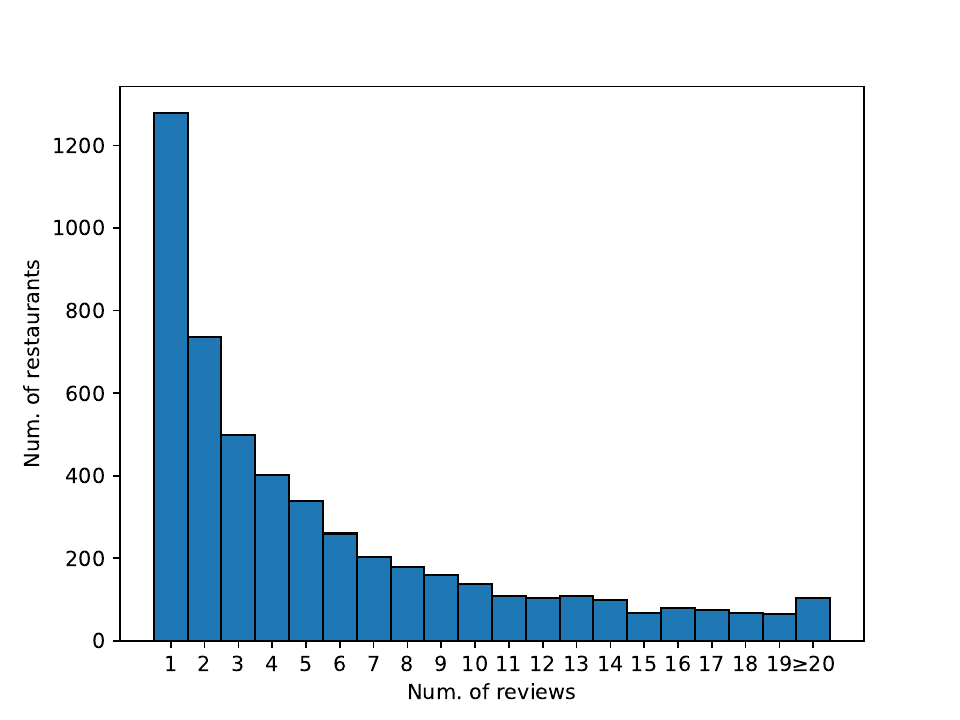}&
        \includegraphics[width=\cw, trim=10 5 40 40, clip]{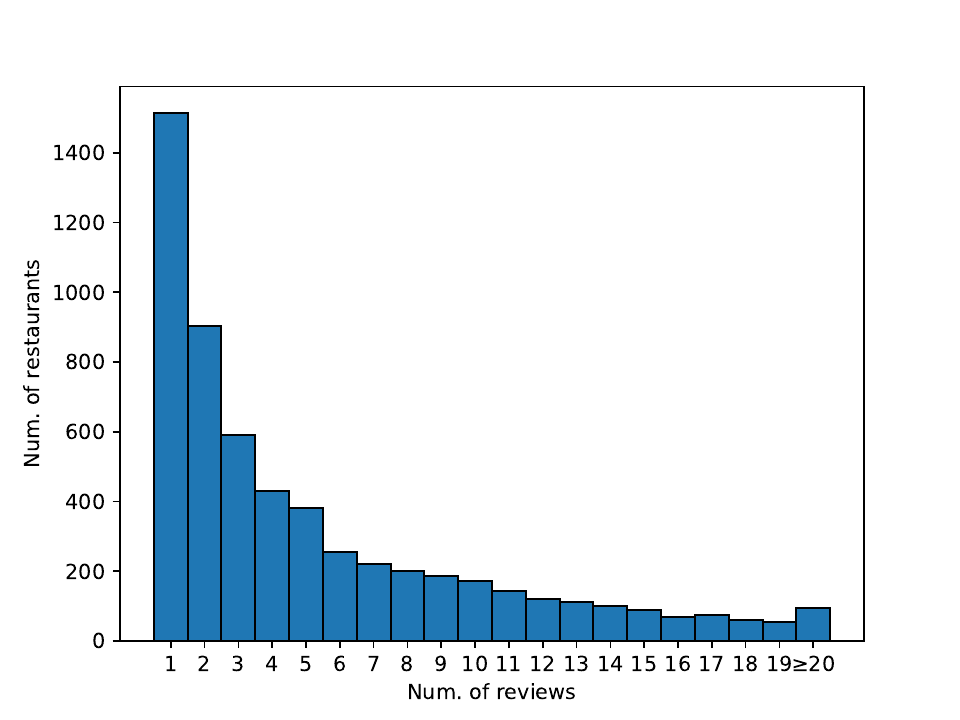}\\
        
        \includegraphics[width=\cw, trim=10 5 40 40, clip]{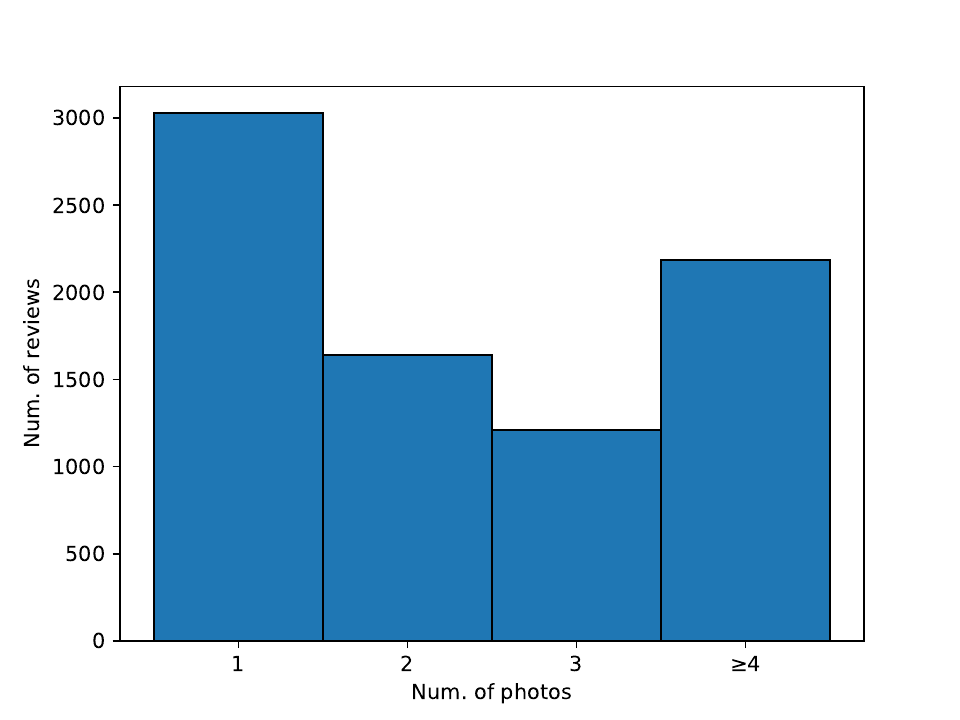}&
        \includegraphics[width=\cw, trim=5 5 40 40, clip]{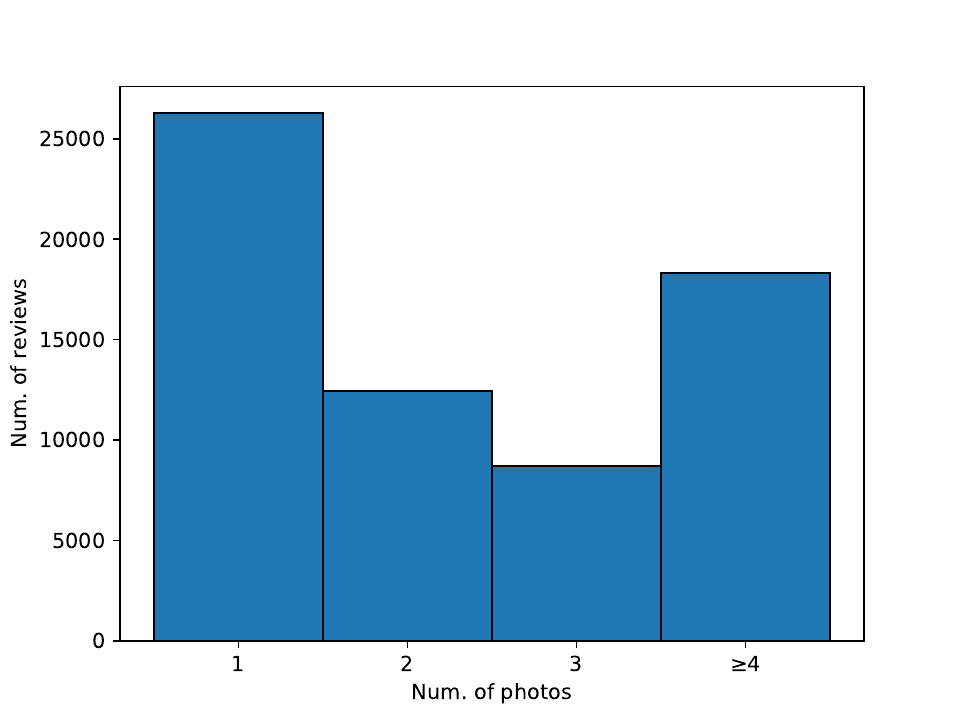}&
        \includegraphics[width=\cw, trim=5 5 40 40, clip]{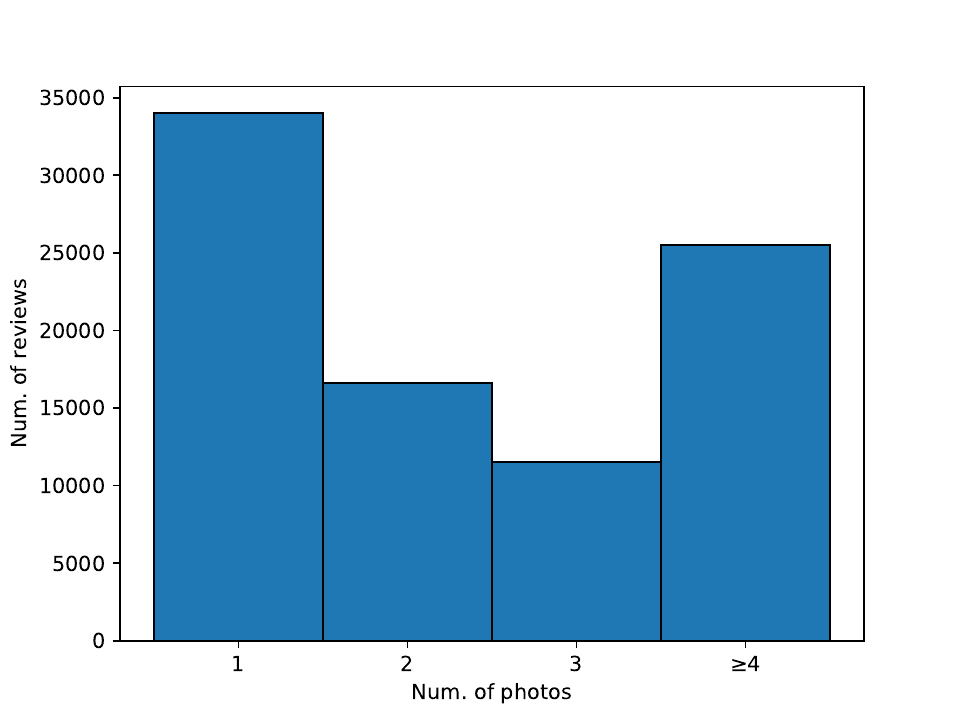}\\
        
        \includegraphics[width=\cw, trim=10 5 40 40, clip]{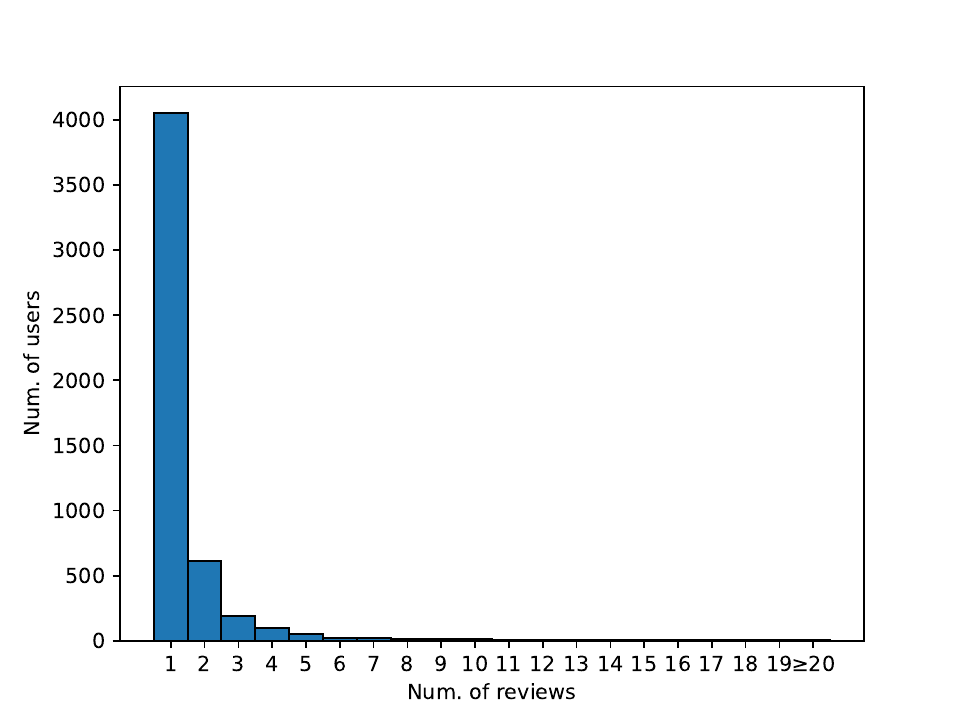}&
        \includegraphics[width=\cw, trim=5 5 40 40, clip]{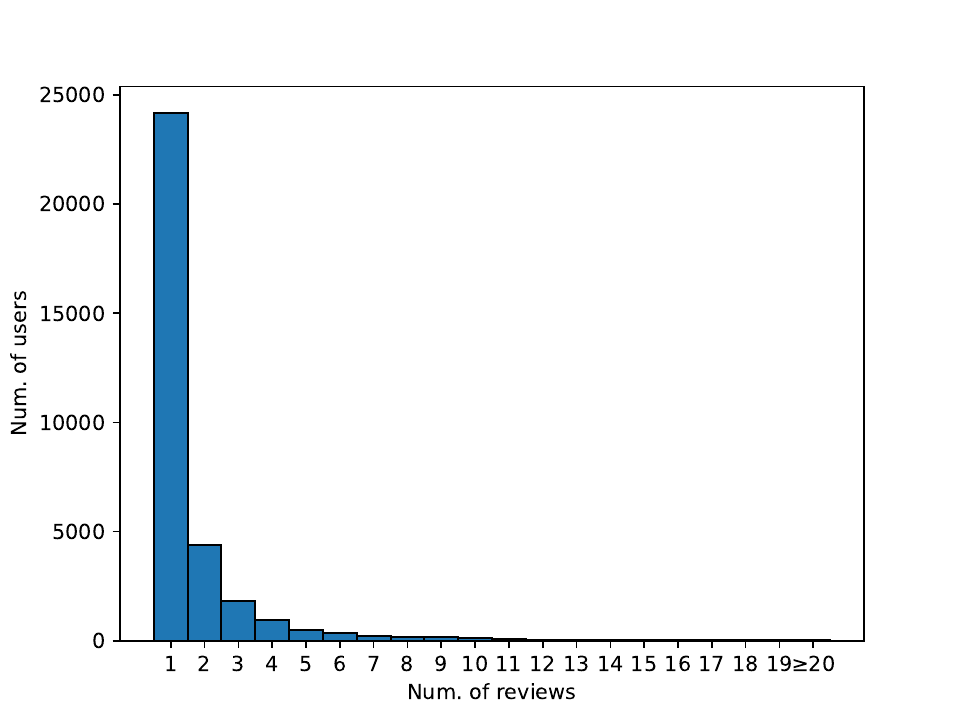}&
        \includegraphics[width=\cw, trim=5 5 40 40, clip]{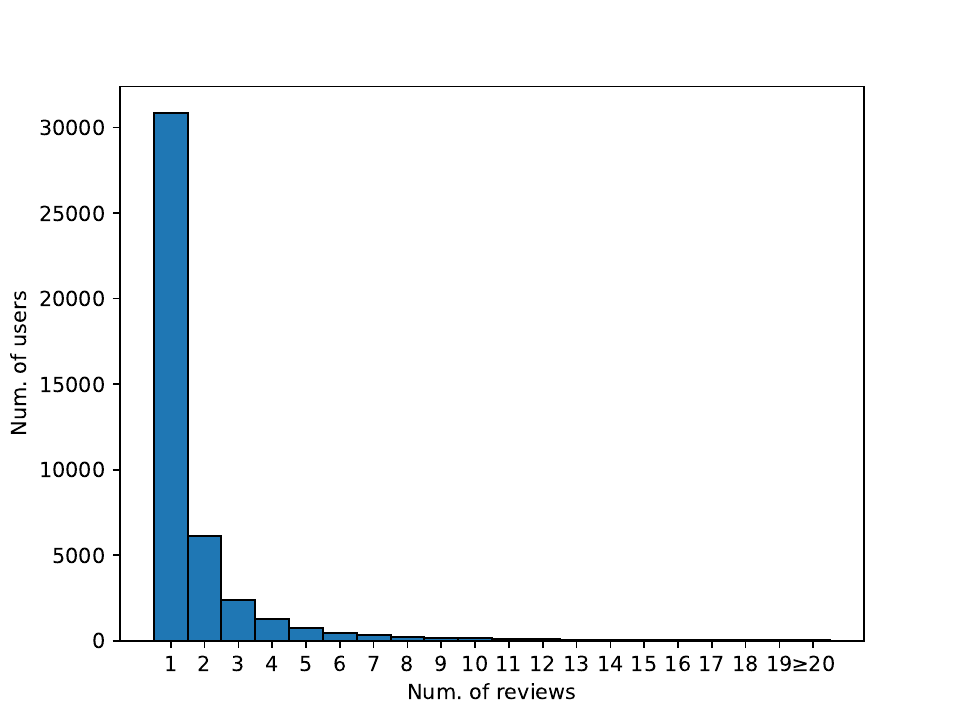}\\
        
        \includegraphics[width=\cw, trim=10 5 40 40, clip]{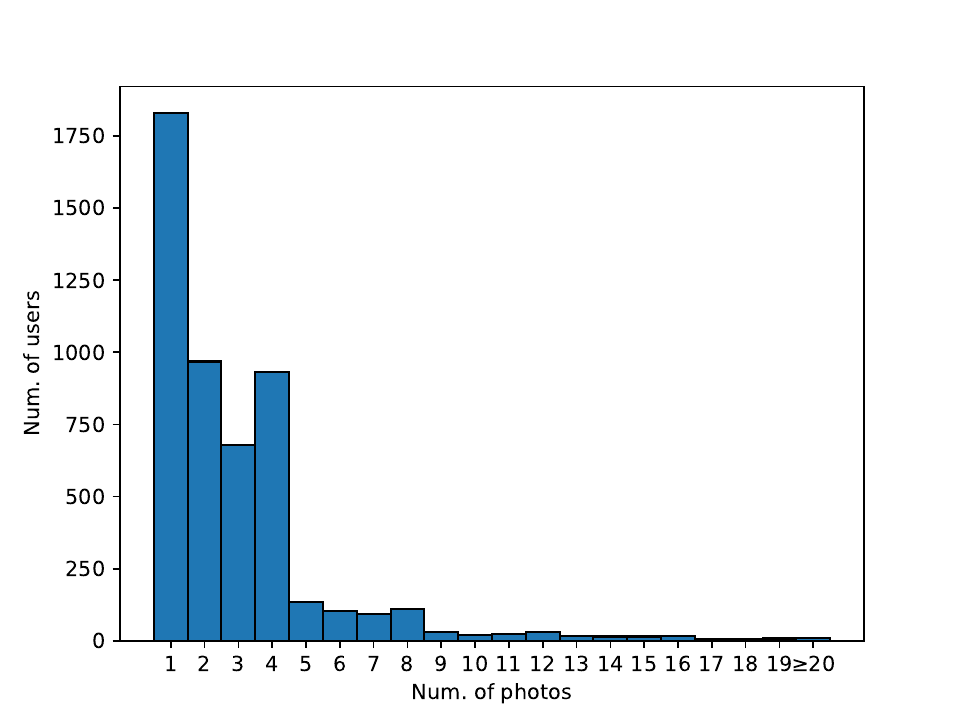}&
        \includegraphics[width=\cw, trim=5 5 40 40, clip]{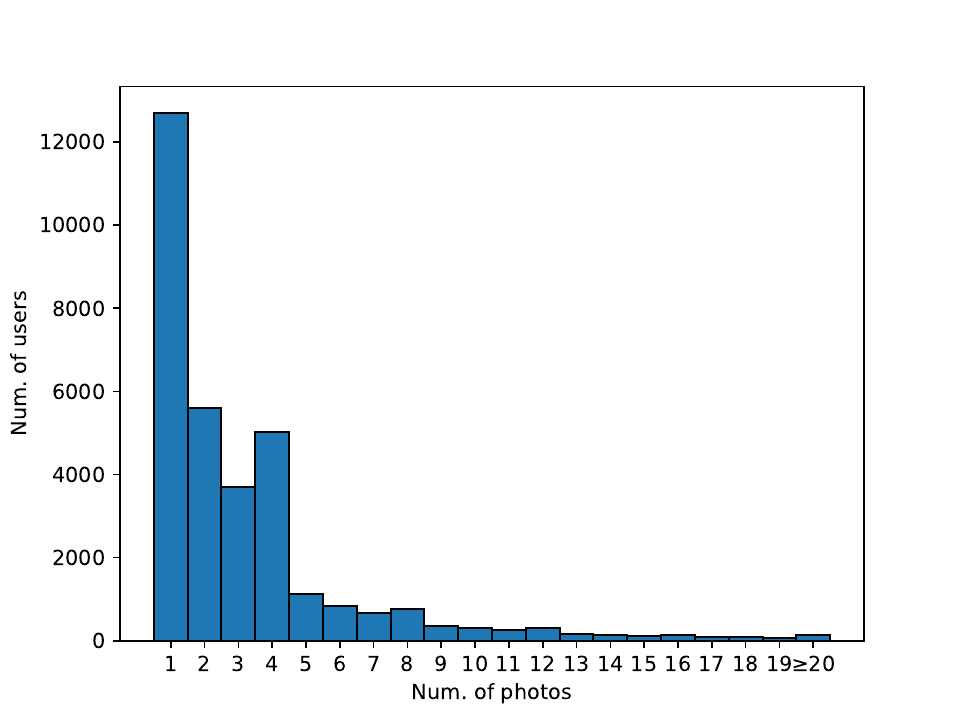}&
        \includegraphics[width=\cw, trim=5 5 40 40, clip]{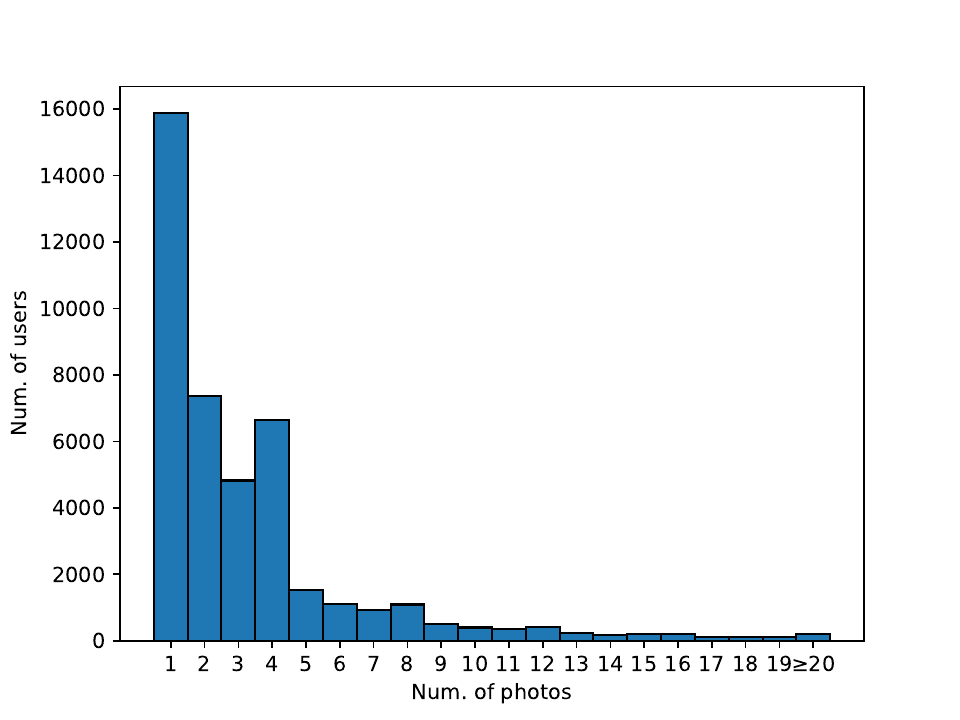}\\
    \end{tabular}
    \caption{Distribution of data downloaded from TripAdvisor (\gijon, \barcelona and \madrid).}
    \label{fig:datasets_fig1}
\end{figure}

\begin{figure}[p]
    \centering
    \begin{tabular}{@{}c@{\espacio}c@{\espacio}c@{}}
        \nyc & \paris & \london \\
        \includegraphics[width=\cw, trim=15 5 40 40, clip]{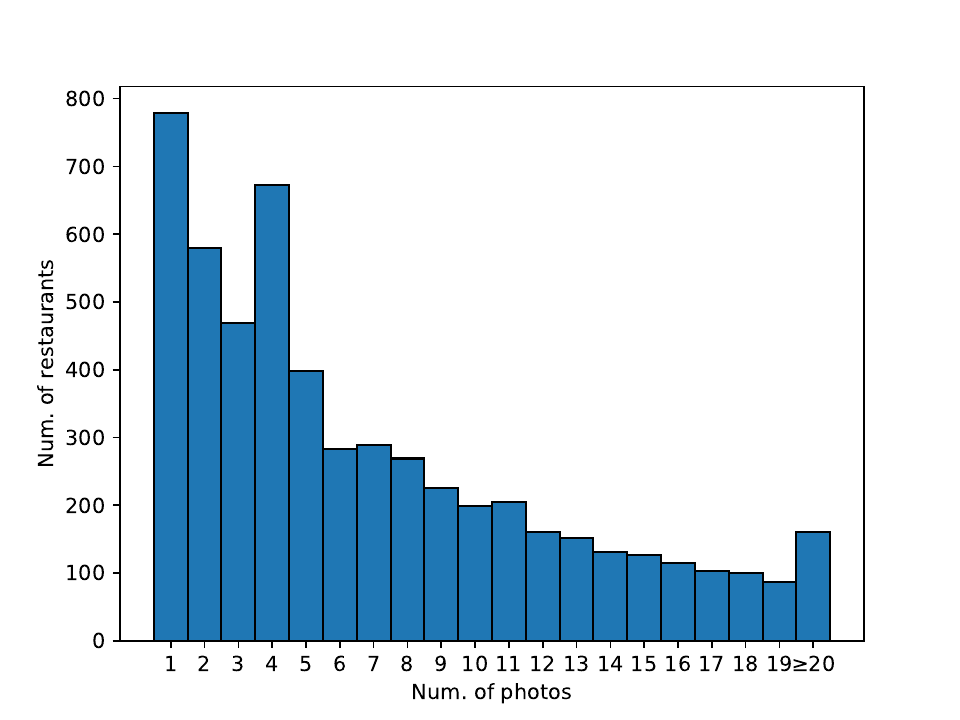}&
        \includegraphics[width=\cw, trim=10 5 40 40, clip]{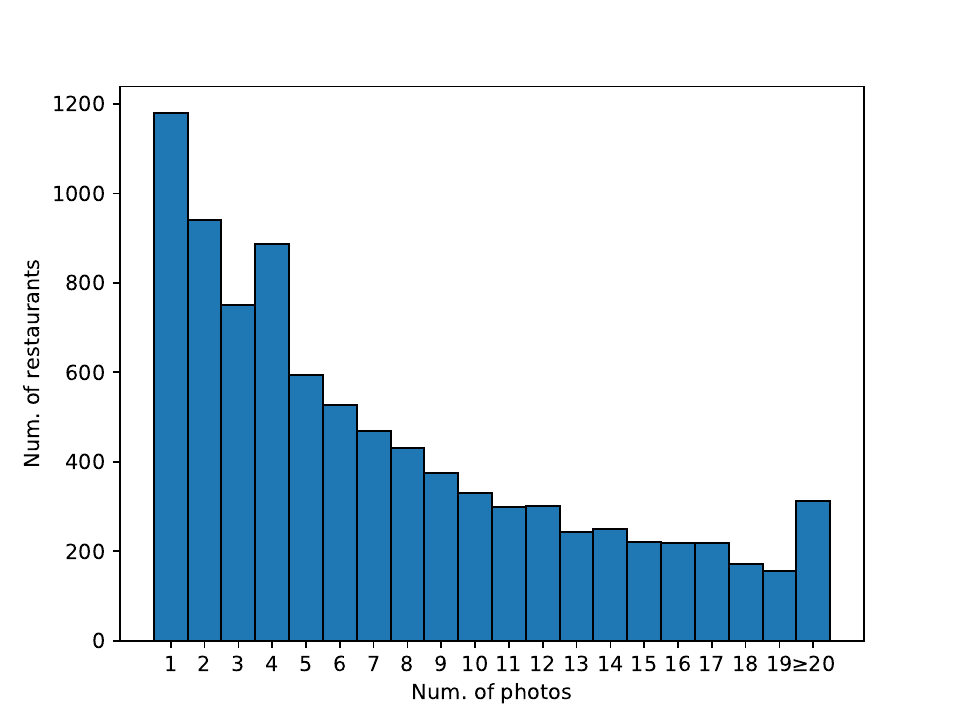}&
        \includegraphics[width=\cw, trim=10 5 40 40, clip]{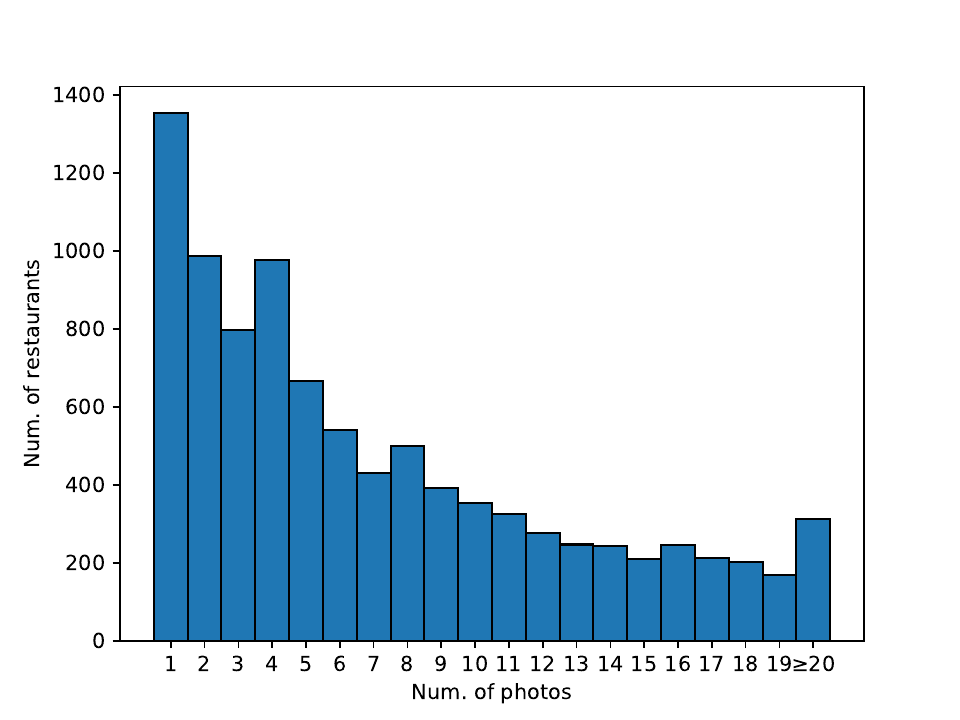}\\
        
        \includegraphics[width=\cw, trim=10 7 40 40, clip]{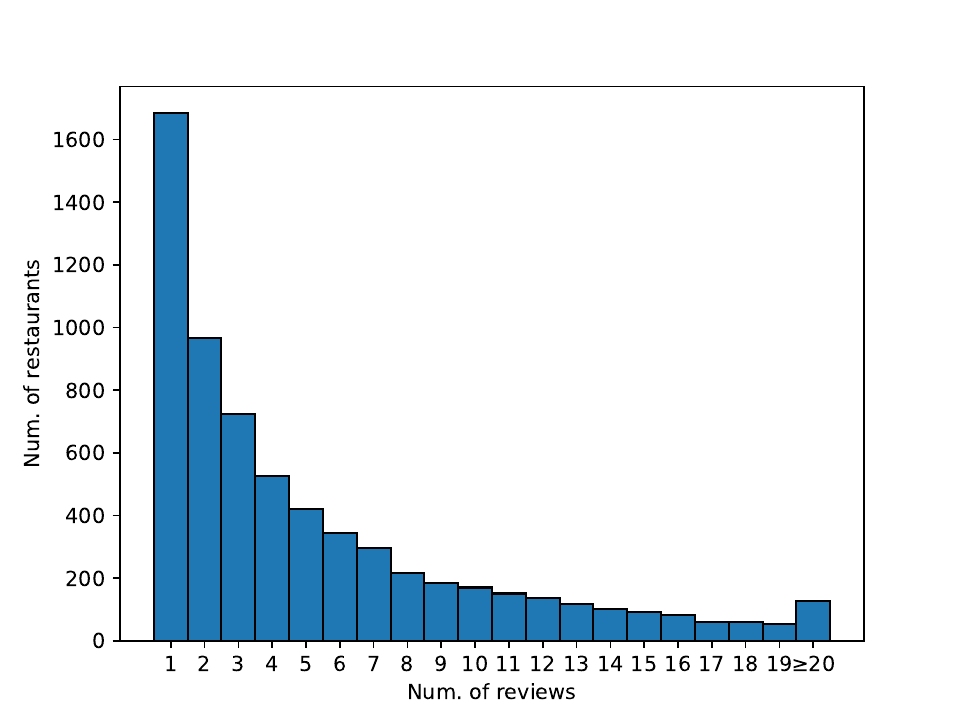}&
        \includegraphics[width=\cw, trim=10 5 40 40, clip]{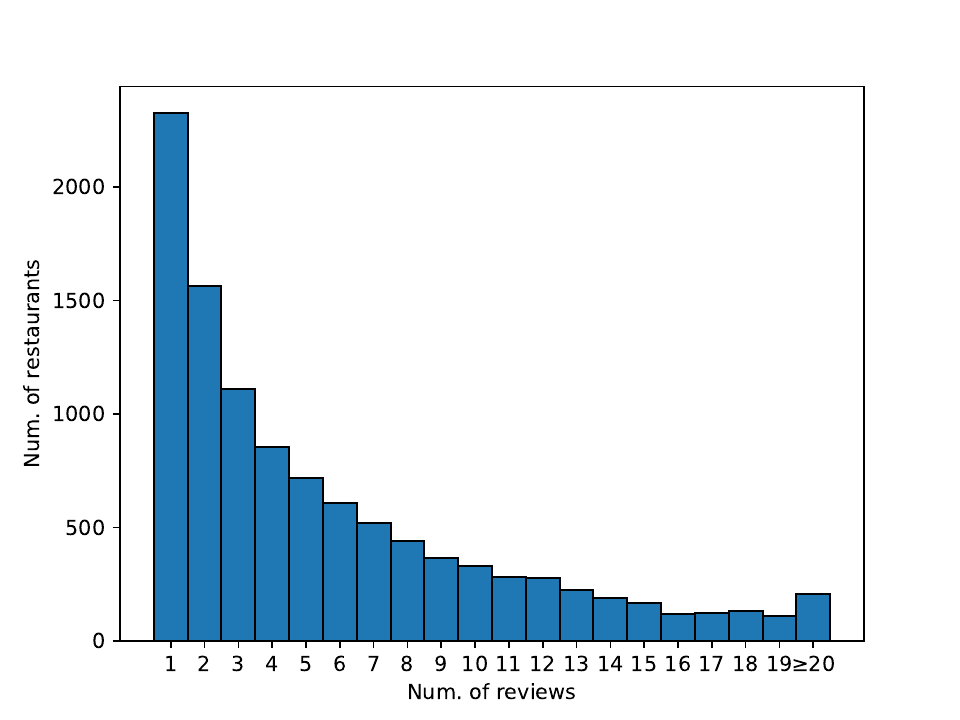}&
        \includegraphics[width=\cw, trim=10 5 40 40, clip]{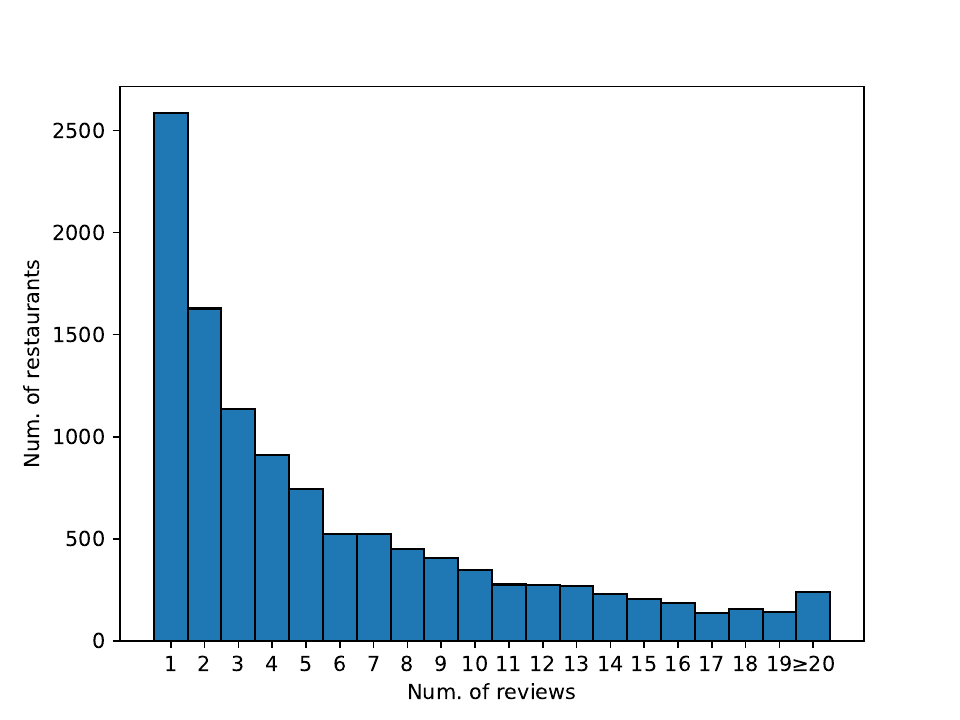}\\
        
        \includegraphics[width=\cw, trim=5 5 40 40, clip]{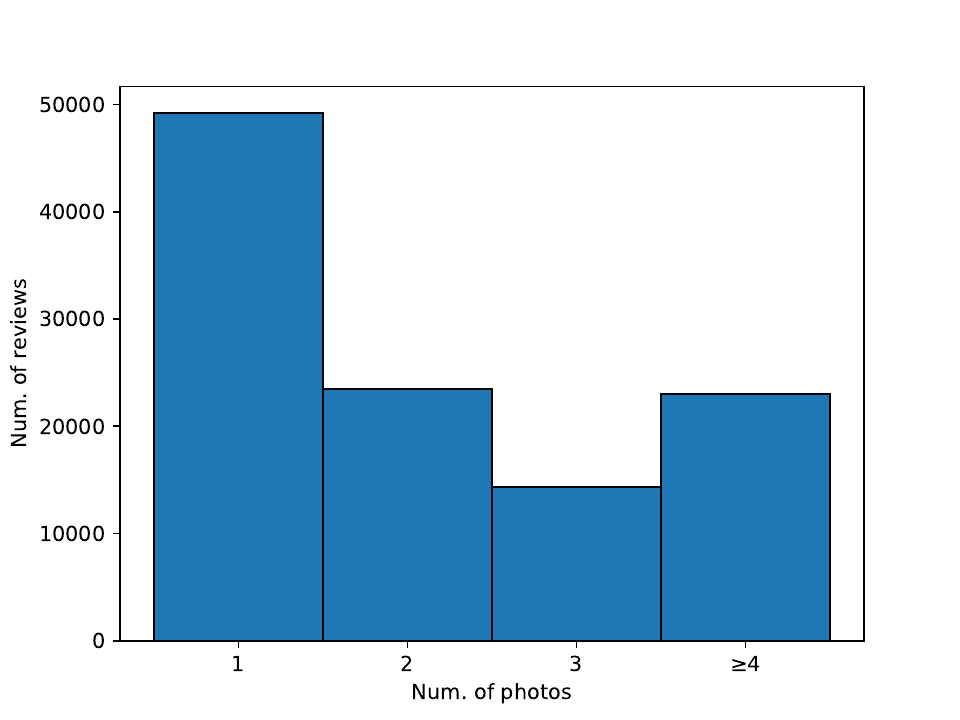}&
        \includegraphics[width=\cw, trim=5 5 40 40, clip]{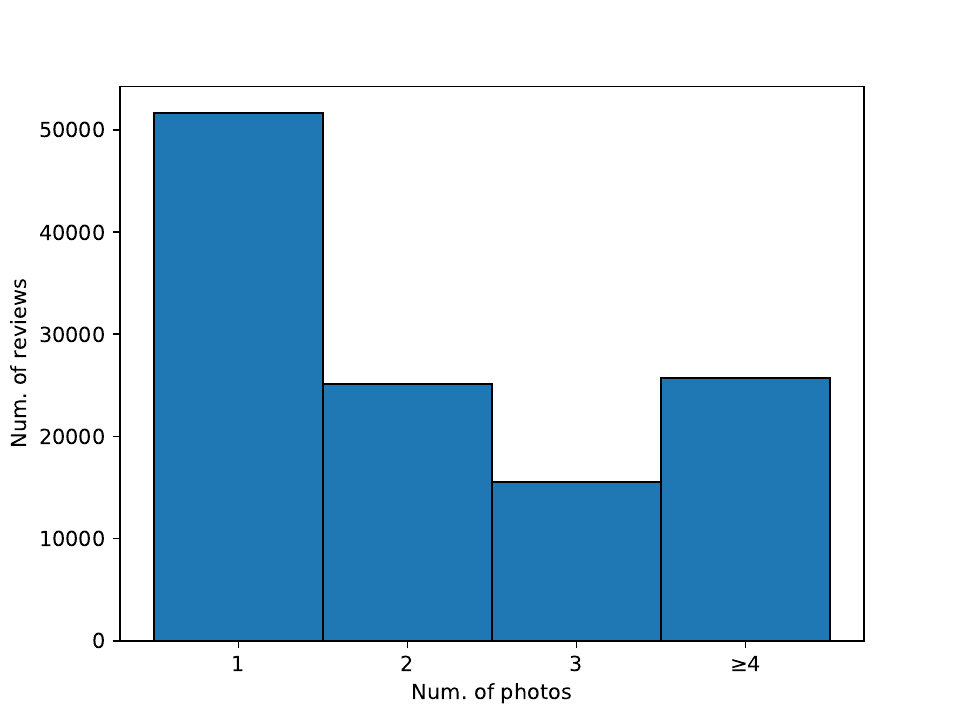}&
        \includegraphics[width=\cw, trim=0 5 40 40, clip]{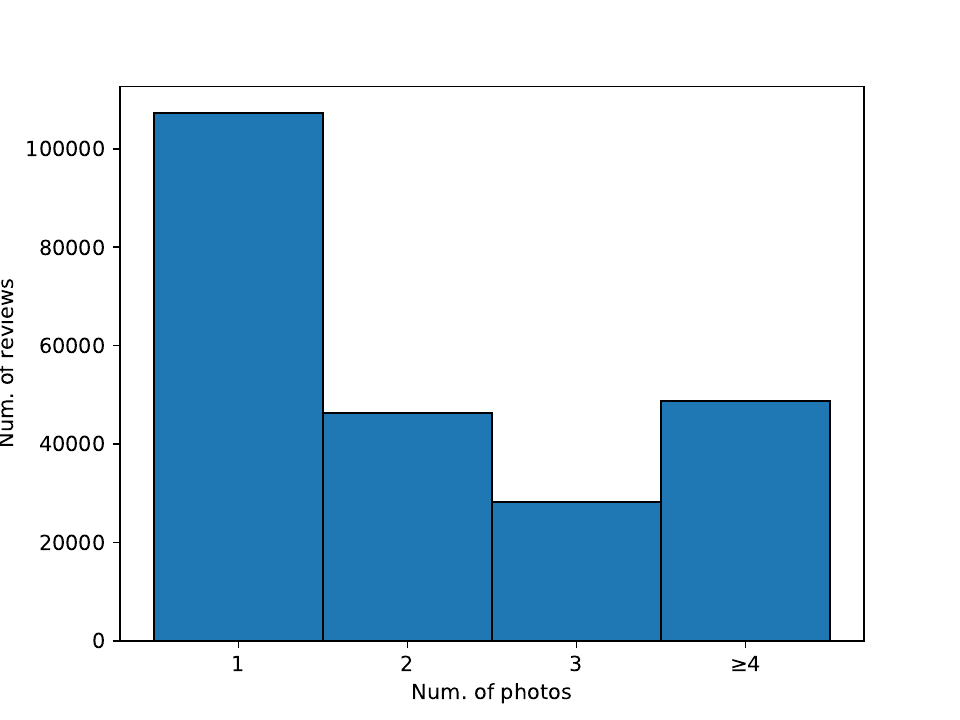}\\
        
        \includegraphics[width=\cw, trim=5 5 40 40, clip]{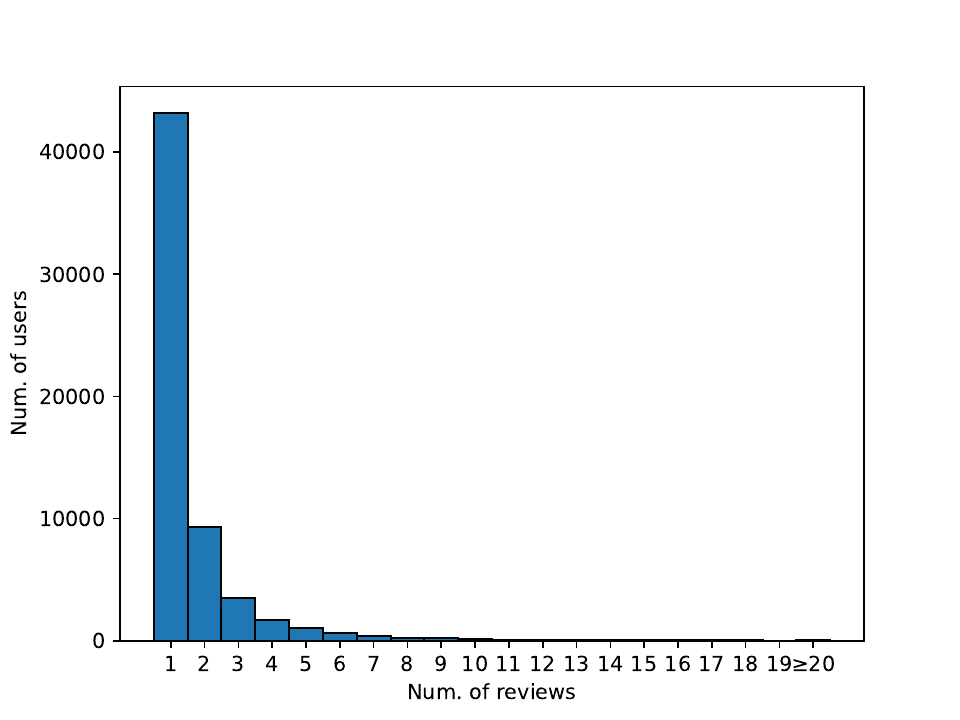}&
        \includegraphics[width=\cw, trim=5 5 40 40, clip]{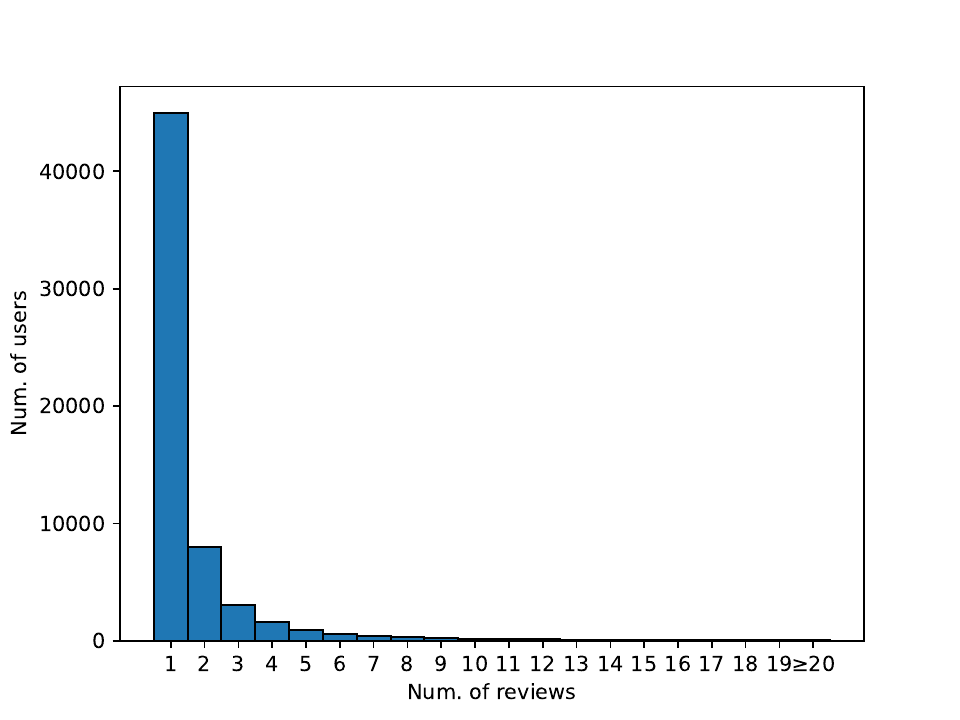}&
        \includegraphics[width=\cw, trim=0 5 40 40, clip]{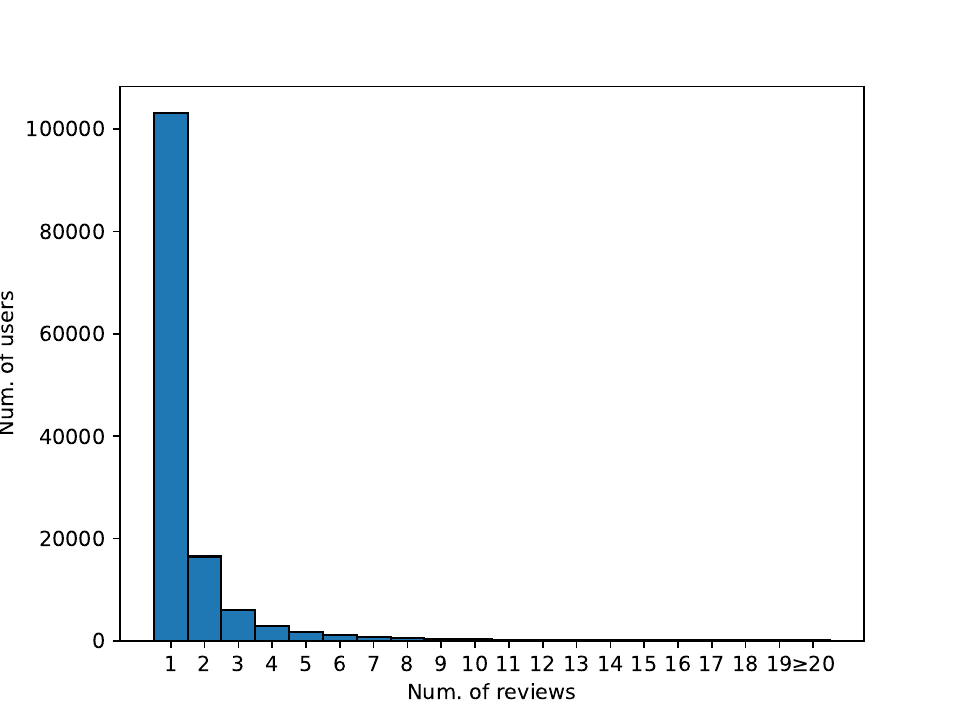}\\
        
        \includegraphics[width=\cw, trim=5 5 40 40, clip]{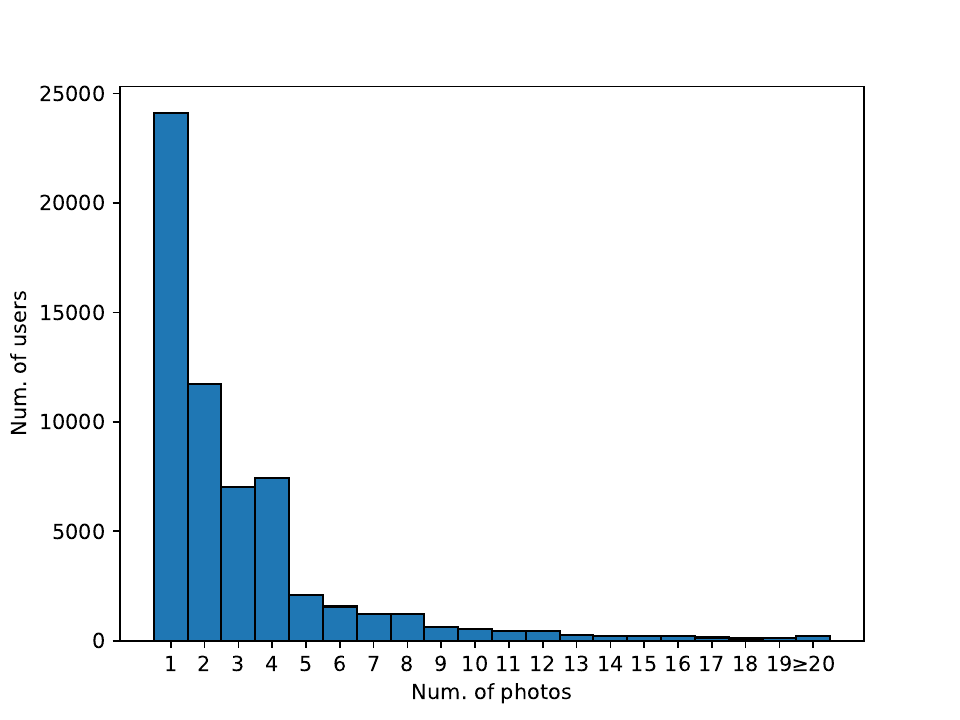}&
        \includegraphics[width=\cw, trim=5 5 40 40, clip]{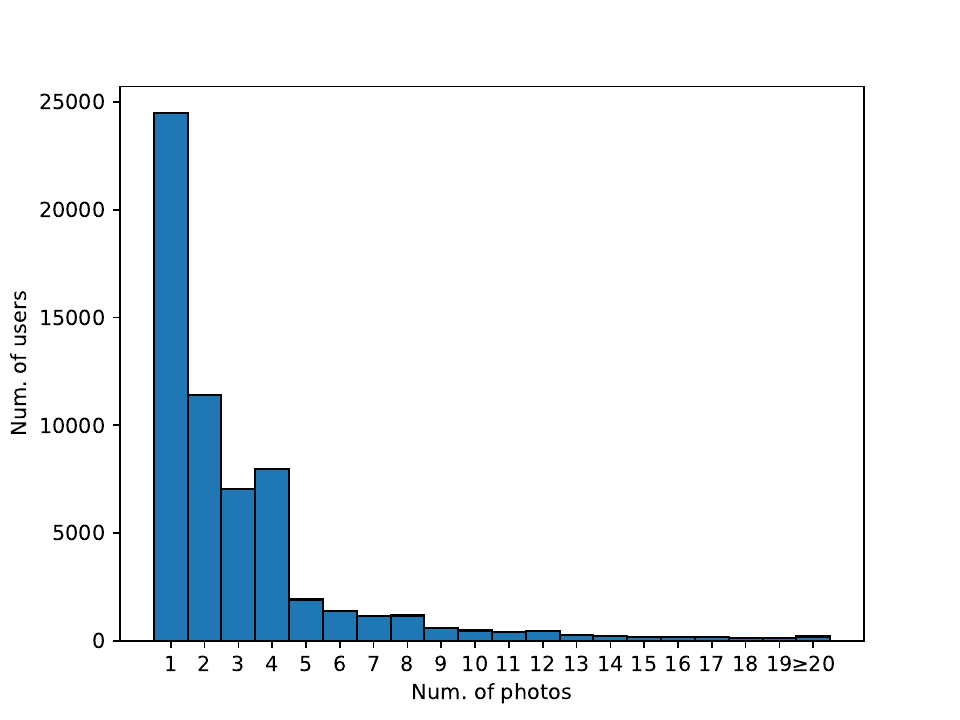}&
        \includegraphics[width=\cw, trim=5 5 40 40, clip]{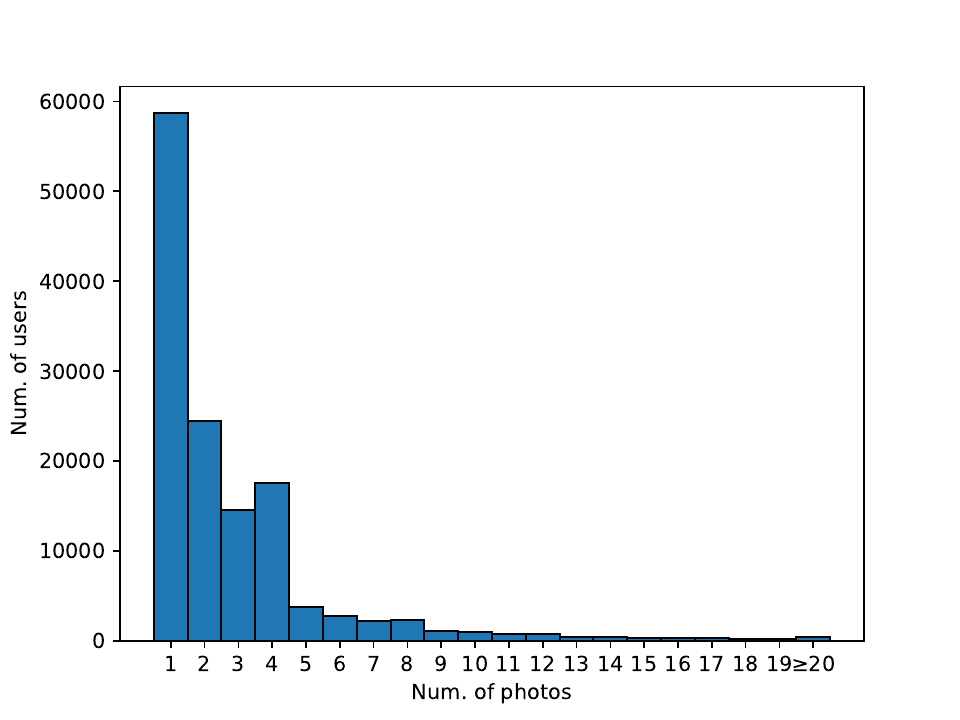}\\
    \end{tabular}
    \caption{Distribution of data downloaded from TripAdvisor (\nyc, \paris and \london).}
    \label{fig:datasets_fig2}
\end{figure}

It is remarkable that the shape of all the distributions mentioned are really similar in the six cities, although the absolute values of the Y axes are quite different, obviously due to the variation in the city sizes.

\subsection{Experiments}
\label{sec:experiments}

In this  section, we describe the comparison details of the approach presented in this paper with two baseline methods devised for this purpose. The first one is called \emph{Random} because its compatibility function,
$\text{Pr}^\mathrm{rnd}(\vec{u}, \vec{f})$,
follows a uniform distribution in $[0, 1]$. RND will stand for this approach in the tables of scores. In order to minimize the implicit bias of the method, we repeat ten times the experiments with this method, and report the average scores so obtained.


The second baseline method is defined by the \emph{Centroid} of the photos available; CNT will stand for this approach. In this way, this method takes into account the codification of the photos used in \oursystem (see Section~\ref{sec:topology}). More precisely, the photos of a restaurant are sorted according to the inverse of the euclidean distance between them and the geometric center (centroid) of the codes of photos of the restaurant. We consider that the most representative photo in a set is the one nearest to the centroid. 

Our implementation of  \oursystem is on Keras \cite{chollet2015keras}, with TensorFlow \cite{martin2016tensorflow} as backend.
It uses the \emph{Adam} \cite{kingma2014adam} optimizer with \emph{linear cosine} \cite{bello2017neural} learning rate decay. Additionally, it is worth noting that the model was trained during 100 epochs.

A grid search was performed looking for the best meta-parameters. For this purpose, we split the training set using again the procedure described in Section~\ref{sec:formal_dataset}. This yields a new training set and a development set. Using these training and development sets, and with a fixed dropout value of $0.2$, we search for the best learning rate in the set \{$5\cdot10^{-3}$, $10^{-3}$, $5\cdot10^{-4}$, $10^{-4}$, $5\cdot10^{-5}$\}. Once the best learning rate is found, we use the original training set to obtain the final model, which is then evaluated in the test set.


\subsection{Evaluation}\label{sec:evaluation}

Let us recall that the model described in Section~\ref{sec:framework} estimates the probability of authorship of a photo by a user, then we can sort the photos available for a restaurant according to this probability. Thus, the evaluation consists of measuring the quality of a ranking.

The methods used in our experiments provide a ranking of a set of photos. To evaluate them we use a top-n framework. In fact, we only need to adapt the measure \cite{cremonesi2010performance} to our context.

As explained in Figure~\ref{fig:train_test_split}, each positive pair $(\vec{u},\vec{f})$ in the test set means that user $\vec{u}$ has taken photo $\vec{f}$, let say in a restaurant $\vec{r}$. Then, we consider as negatives all pairs formed by the same user $\vec{u}$ and all the photos of the same restaurant $\vec{r}$, but taken by users other than $\vec{u}$ (generically represented as $\vec{g}$). In symbols,
\begin{equation}
\begin{split}
    &\vec{r} \in \itemsset, \vec{u} \in \usersset,\\
    &\vec{f} \in \operatorname{photos}(\Vec{u},\vec{r}),\\
    &\vec{g} \in \operatorname{photos}(\vec{r}) \setminus \text{photos}(\vec{u}).    
\end{split}\label{eq:test_quiz}
\end{equation}

The objective is that the method will be able to place the user's picture, $\vec{f}$, in the first position of the ranking of all these photos when ordered by the joint (authorship) probability, $\Pr(\vec{u}, \vec{f})$.
In the following sections, we report two types of measures to assess this ranking.

First, we count the number of times that each method places the user's photo, $\vec{f}$, among the top $n$ positions. Taking into account how we devised the evaluation process, in which there is only one correct photo to be ranked among others in the highest possible position, this top-n measure coincides with \emph{Recall at n}, and it is proportional to \emph{Precision at n}. More specifically, it is equivalent to $n \times \mathit{Precision}@n$, expressed as a percentage. \citet{Herlocker2004} claim that these measures are adequate for tasks of the type \emph{Find Good Items}, which is our case.

However, the top-n measure is optimistically biased when the number of photos to be ordered is lower than the value of $n$.  Obviously, any ranking with less than $n$ photos will be considered as a successful top-n prediction, provided it will always contain the correct photo. Thus, we will also analyze the quality of the rankings with a second measure that will take into account their variable length. This measure is the percentile position of the correct photo in the ranking, which will be computed as

\begin{equation}
    \operatorname{percentile}(\vec{f},\vec{R})=100\cdot\frac{\operatorname{index}(\vec{f},\vec{R})-1}{|\vec{R}|},
\end{equation}
where $\vec{R}=\{\vec{f}\} \cup (\operatorname{photos}(\vec{r}) \setminus \text{photos}(\vec{u})) $ is the ranking containing the photo $\vec{f}$ taken by user $\vec{u}$ together with all the photos of the same restaurant but taken by other users. Here, $\operatorname{index}(\vec{f},\vec{r})$ is the position of $\vec{f}$ in the ranking.

The ranking is in descendant order of $\Pr(\vec{u}, \vec{f})$; therefore, the lower the percentile, the better the ranking.

The way in which this measure takes into account the variable length of rankings can be illustrated with the following example: let's suppose a ranking with only two photos where $\vec{f}$ is in second position. This situation yields a percentile value of 50\%. The same second position for another ranking with 100 photos will yield a value of 1\%. This difference reflects that being the second of two is worse than being the second of 100, which seems reasonable.

\subsubsection{Performance analysis in terms of users' satisfaction}
\label{sec:satisfaction}

The most obvious way to know the users' satisfaction with different ranking approaches would be to perform some kind of surveys regarding their happiness with the predictions. \citet{Herlocker2004} mention this approach as \emph{explicit} evaluation. However, surveys are difficult to carry out, mainly because we need to interact with the users through the service provider (in our case via the TripAdvisor site), so we devised the \emph{implicit} (in terms of \citep{Herlocker2004}) evaluation approach explained in Section~\ref{sec:evaluation}, which goes beyond computing accuracy and ``judge the quality of recommendations as users see them: as recommendation lists'' \citep{McNee2006}. 

We are making a reasonable assumption with this evaluation procedure: if a method is able to rank a given user's photo in top positions when mixed with other users' photos it is because it was able to capture (to some extent) the essence of the authorship of photos. Therefore, the top ranked photos predicted for a user will mostly share such essence and we hopefully expect them to satisfy the user. In other words, we assume the top-n score as a user satisfaction measure.

\begin{figure*}[tb]
    \centering
    \includegraphics[width=\textwidth]{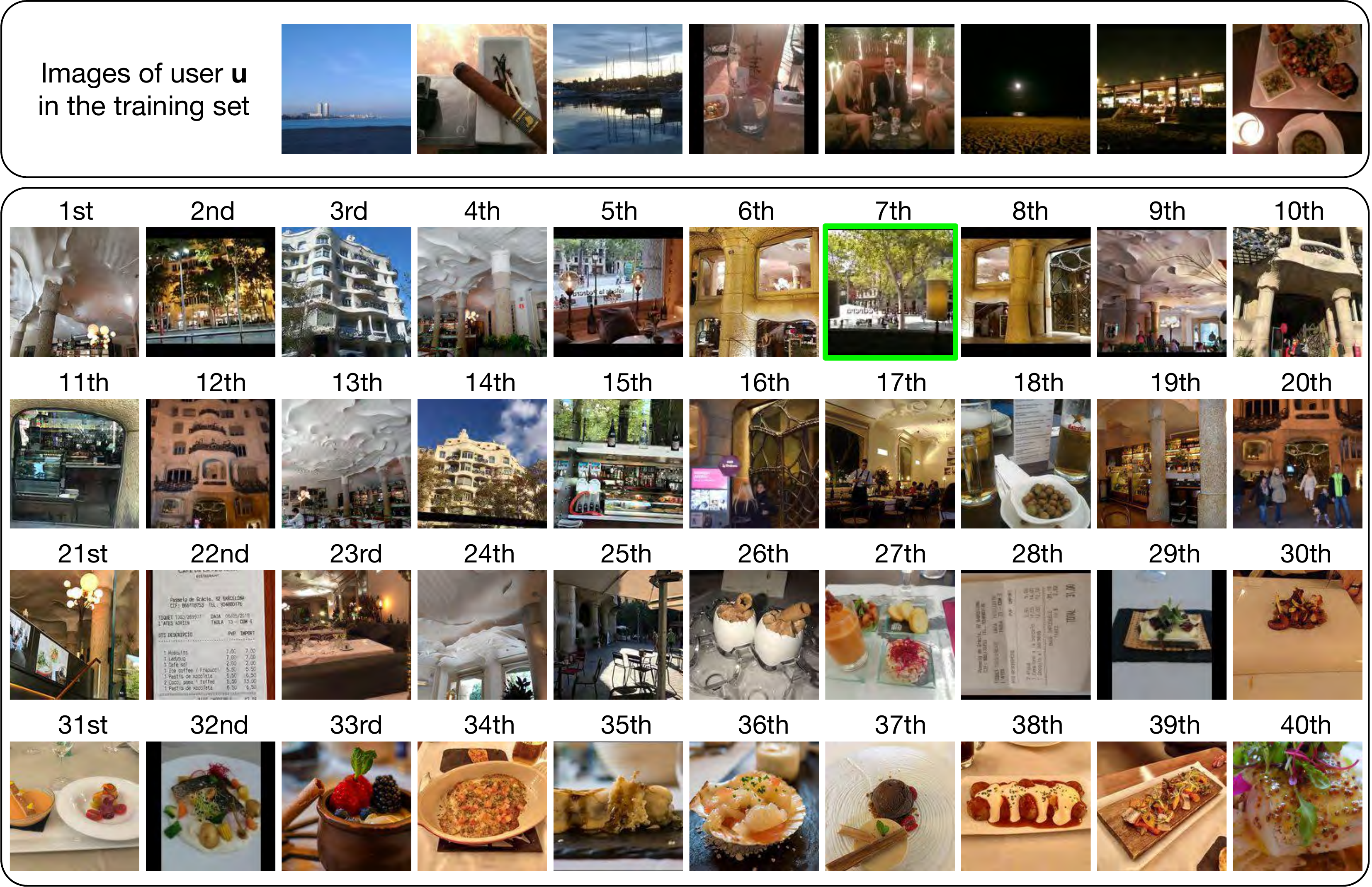}\vspace*{1em}
    \caption{The first row shows the eight photos in the training set taken by a user \vec{u}, while the bottom box exhibits the photos of a restaurant of the test set ordered (left to right) by the joint probability estimated by \oursystem. We observe that the user really likes photos of the area but not of food, and those are the pictures placed in the first positions. The seventh photo was really taken by the user in this test restaurant. Note that some photos were slightly retouched in size and/or luminosity for a correct visualization in this paper}\label{fig:ordering}
\end{figure*}

Let us show an example to illustrate this idea. The first row of Figure~\ref{fig:ordering} shows the eight photos taken by a user included in the training set. As can be appreciated, only one out of eight photos shows food, while the remainder ones show mostly exterior or interior areas of the restaurant, but not food. Once the joint probability, $\Pr(\vec{u}, \vec{f})$, was learned by \oursystem, we tested the model with all the photos of a restaurant that was not used for training purposes for this user (only one of those photos was taken by the user that we are analyzing). The 40 photos of this restaurant were ordered according to user's preferences. The ranking obtained is reproduced in the bottom box of Figure~\ref{fig:ordering}. It is worth noting that we can only find food in the photos  ranked in the last positions.

Regarding the photo taken by the user in this restaurant, it is ranked in the seventh position by \oursystem. Notice also that none of the photos ranked on top of the true one are food, they are from the exterior/interior of the restaurant, just like most of the photos taken by the user in other restaurants. The peculiarity of the user was grasped by \oursystem.

To check the performance of \oursystem we carried out evaluation experiments on test sets of the six cities. Table~\ref{tab:top_n_scores} shows the percentage of test examples ranked in top positions for the six cities. We filtered out restaurants with less than 10 photos in the test in order to avoid the optimistic bias (see Section~\ref{sec:evaluation}) of the top-n measure, and users with less than 10 photos in the training to ensure a reasonable amount of information to learn from.

The scores are quite similar in all the cities, and there are big differences between the three methods. The baseline CNT (\emph{centroid}) is the worst in every row of the table, which means that the vectorial representation of photos obtained by the CNN is not guided by any semantic rule related to the users' taste. Looking at the top-10 scores, \oursystem is around 20 percentage points better than RND (\emph{random}) in all cities except \gijon, where we only achieved around 12\% of improvement over RND. This is mainly because of the smaller size of its dataset. The best scores were obtained in the data from \paris, the second largest dataset used in our experiments.


\begin{table}[tb]
    \begin{center}\small
        \begin{tabular}{c@{\espacio}|@{\espacio}rrr@{\espacio}|@{\espacio}rrr@{\espacio}|@{\espacio}rrr}
            \toprule 
             \small
                & \multicolumn{3}{c@{\espacio}|@{\espacio}}{\gijon (338)} &\multicolumn{3}{c@{\espacio}|@{\espacio}}{\barcelona (3023)} &\multicolumn{3}{c}{\madrid (4578)}\\
                TOP & RND & CNT & \oursystem & RND & CNT & \oursystem & RND & CNT & \oursystem \\
                \midrule
                1 & 4.3\% & 2.7\% & 8.9\% & 4.0\% & 1.6\% & 11.8\% & 3.6\% & 1.6\% & 11.9\% \\ 
                2 & 8.3\% & 5.9\% & 16.3\% & 7.9\% & 4.0\% & 22.2\% & 7.5\% & 3.6\% & 20.3\% \\ 
                3 & 11.8\% & 7.7\% & 20.1\% & 11.8\% & 6.1\% & 29.3\% & 11.3\% & 5.9\% & 27.9\% \\ 
                4 & 15.7\% & 11.2\% & 26.6\% & 15.8\% & 9.2\% & 34.9\% & 14.9\% & 8.8\% & 33.5\% \\ 
                5 & 19.6\% & 15.7\% & 29.9\% & 19.9\% & 12.2\% & 39.8\% & 18.7\% & 11.9\% & 38.8\% \\ 
                6 & 23.7\% & 18.9\% & 35.2\% & 23.7\% & 15.9\% & 44.9\% & 22.4\% & 15.1\% & 43.2\% \\ 
                7 & 27.9\% & 21.6\% & 40.2\% & 27.7\% & 20.1\% & 49.0\% & 26.1\% & 18.5\% & 47.0\% \\ 
                8 & 32.5\% & 24.0\% & 42.9\% & 31.8\% & 23.6\% & 53.0\% & 29.9\% & 22.4\% & 50.6\% \\ 
                9 & 36.9\% & 27.2\% & 46.7\% & 35.9\% & 27.9\% & 56.3\% & 33.6\% & 26.5\% & 53.8\% \\ 
                10 & 40.9\% & 35.5\% & 52.1\% & 39.9\% & 32.8\% & 59.7\% & 37.3\% & 31.1\% & 57.2\% \\ 
                \bottomrule
        \end{tabular}
        
        \begin{tabular}{c@{\espacio}|@{\espacio}rrr@{\espacio}|@{\espacio}rrr@{\espacio}|@{\espacio}rrr}
            \toprule 
             \small
                & \multicolumn{3}{c@{\espacio}|@{\espacio}}{\nyc (4230)} &\multicolumn{3}{c@{\espacio}|@{\espacio}}{\paris (4625)} &\multicolumn{3}{c}{\london (9176)}\\
                TOP & RND & CNT & \oursystem & RND & CNT & \oursystem & RND & CNT & \oursystem \\
                \midrule
                1 & 3.8\% & 1.6\% & 11.6\% & 4.6\% & 1.9\% & 13.9\% & 3.4\% & 1.7\% & 11.5\% \\
                2 & 7.4\% & 3.7\% & 20.1\% & 9.3\% & 4.3\% & 22.5\% & 6.9\% & 3.5\% & 19.3\% \\
                3 & 11.2\% & 5.6\% & 26.9\% & 13.8\% & 6.9\% & 29.7\% & 10.3\% & 5.4\% & 25.5\% \\
                4 & 14.9\% & 8.0\% & 32.4\% & 18.3\% & 10.3\% & 35.8\% & 13.7\% & 7.8\% & 30.9\% \\
                5 & 18.6\% & 11.3\% & 36.8\% & 22.8\% & 13.5\% & 42.0\% & 17.1\% & 10.6\% & 35.7\% \\
                6 & 22.3\% & 14.2\% & 41.4\% & 27.3\% & 17.4\% & 47.6\% & 20.5\% & 13.9\% & 39.9\% \\
                7 & 26.0\% & 18.3\% & 45.3\% & 31.9\% & 22.6\% & 52.3\% & 24.0\% & 17.2\% & 43.8\% \\
                8 & 29.7\% & 22.3\% & 49.2\% & 36.4\% & 27.4\% & 56.5\% & 27.5\% & 20.5\% & 47.4\% \\
                9 & 33.5\% & 26.2\% & 52.4\% & 40.8\% & 33.1\% & 60.4\% & 30.9\% & 24.3\% & 50.1\% \\
                10 & 37.3\% & 30.1\% & 55.3\% & 45.2\% & 39.3\% & 64.4\% & 34.2\% & 28.2\% & 53.1\% \\
                \bottomrule
        \end{tabular}
        
        \caption{Percentage of test cases in top-n positions in the six cities (the larger, the better). The values in parentheses are the number of test cases considered for this experiment after filtering out those restaurants and users with less than 10 photos in the training and test sets, respectively}
        \label{tab:top_n_scores}
    \end{center}
\end{table}

\subsubsection{Performance analysis regarding the amount of users' information}
\label{sec:percentiles}

\begin{figure}[p]
    \newcommand{\factor}{1}
    \includegraphics[width=\factor\linewidth]{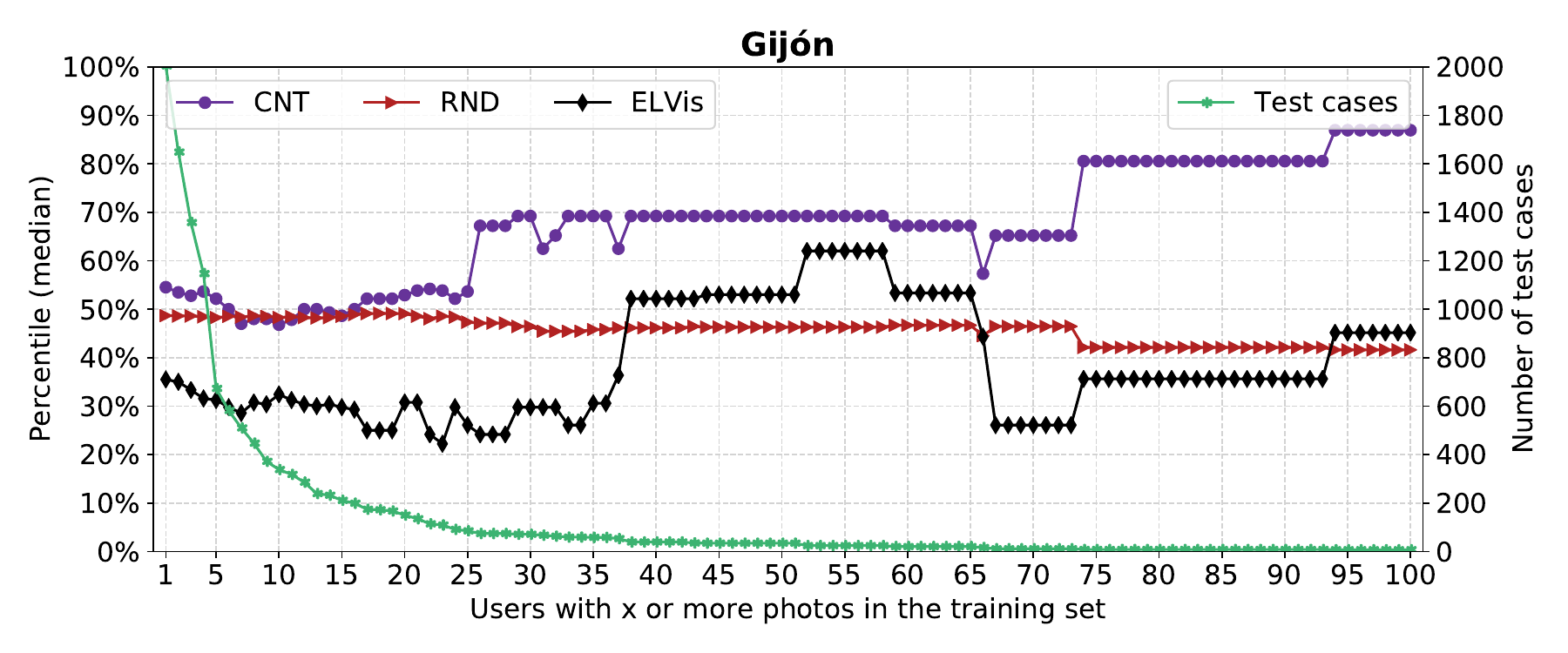}\\
    \includegraphics[width=\factor\linewidth]{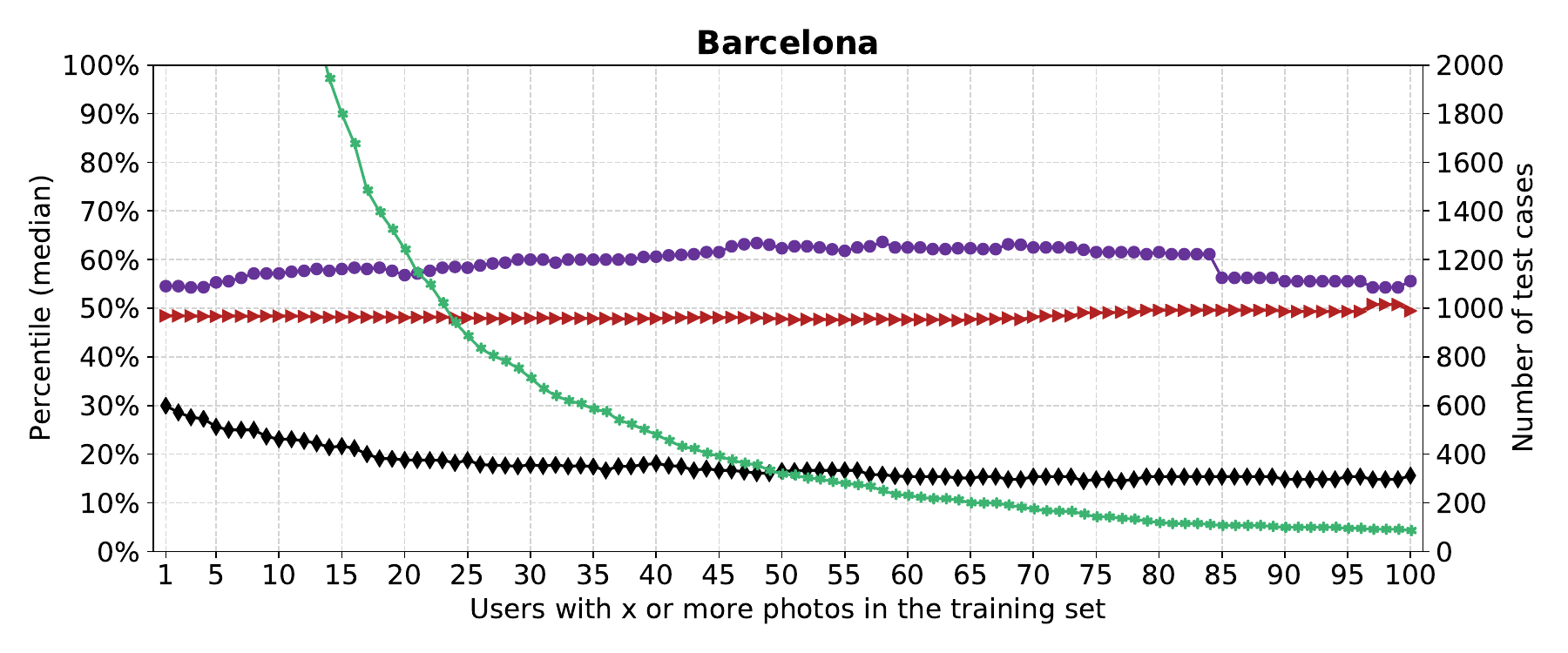}\\
    \includegraphics[width=\factor\linewidth]{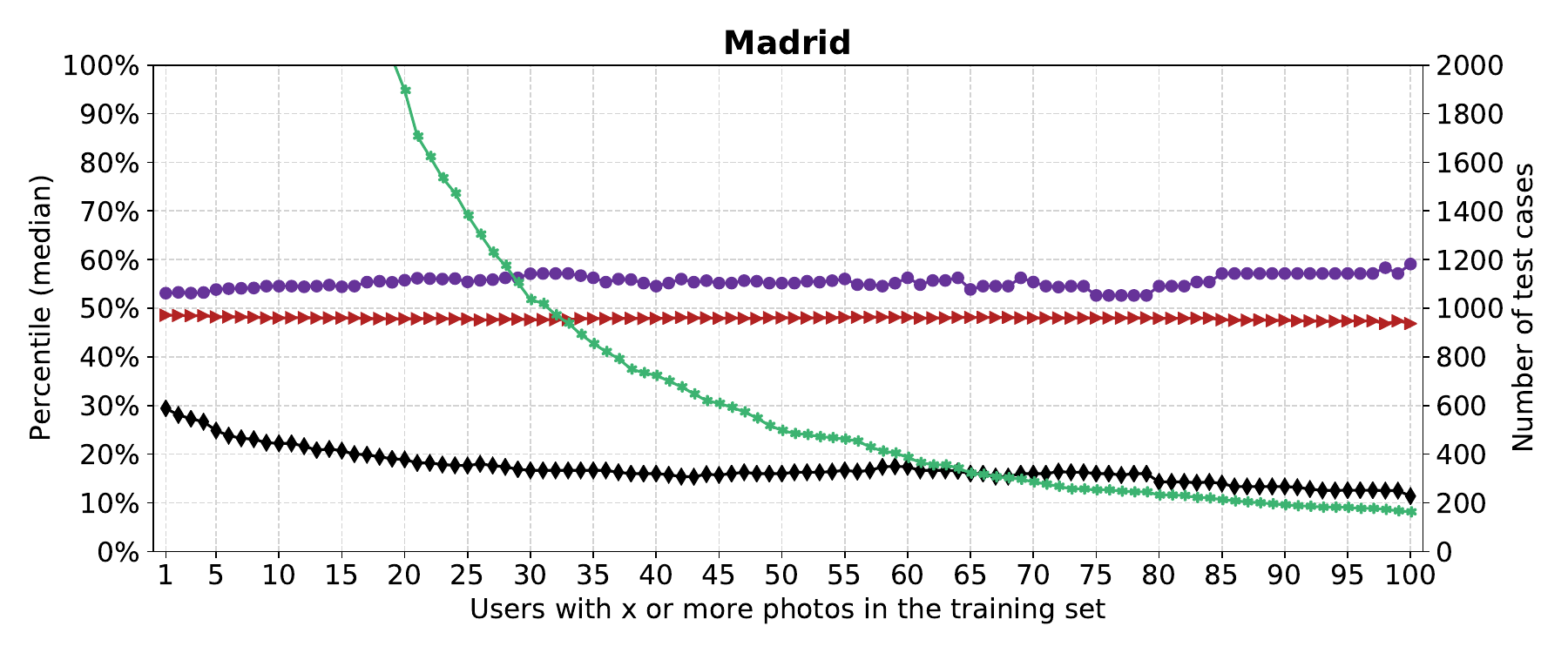}
    \caption{Average percentiles of the test photos, in the three Spanish cities, for users with different number of photos in training. The green line represents the number of cases available below 2,000 (see the vertical axes in the right side). We appreciate that the lack of {test cases} in \gijon leads to an irregular behavior of \oursystem for users with 35 photos or more in the training set}
    \label{fig:results}
\end{figure}

\begin{figure}
    \newcommand{\factor}{1}
    \includegraphics[width=\factor\linewidth]{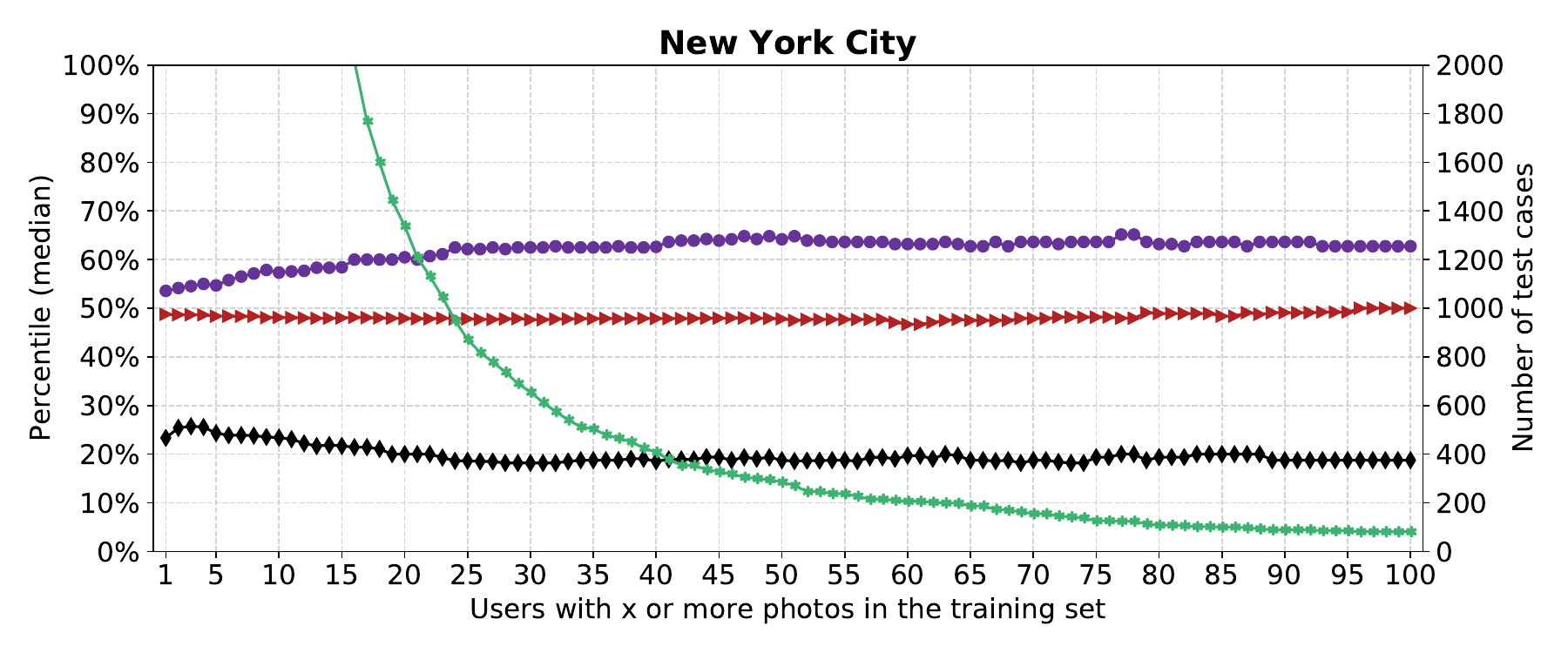}\\
    \includegraphics[width=\factor\linewidth]{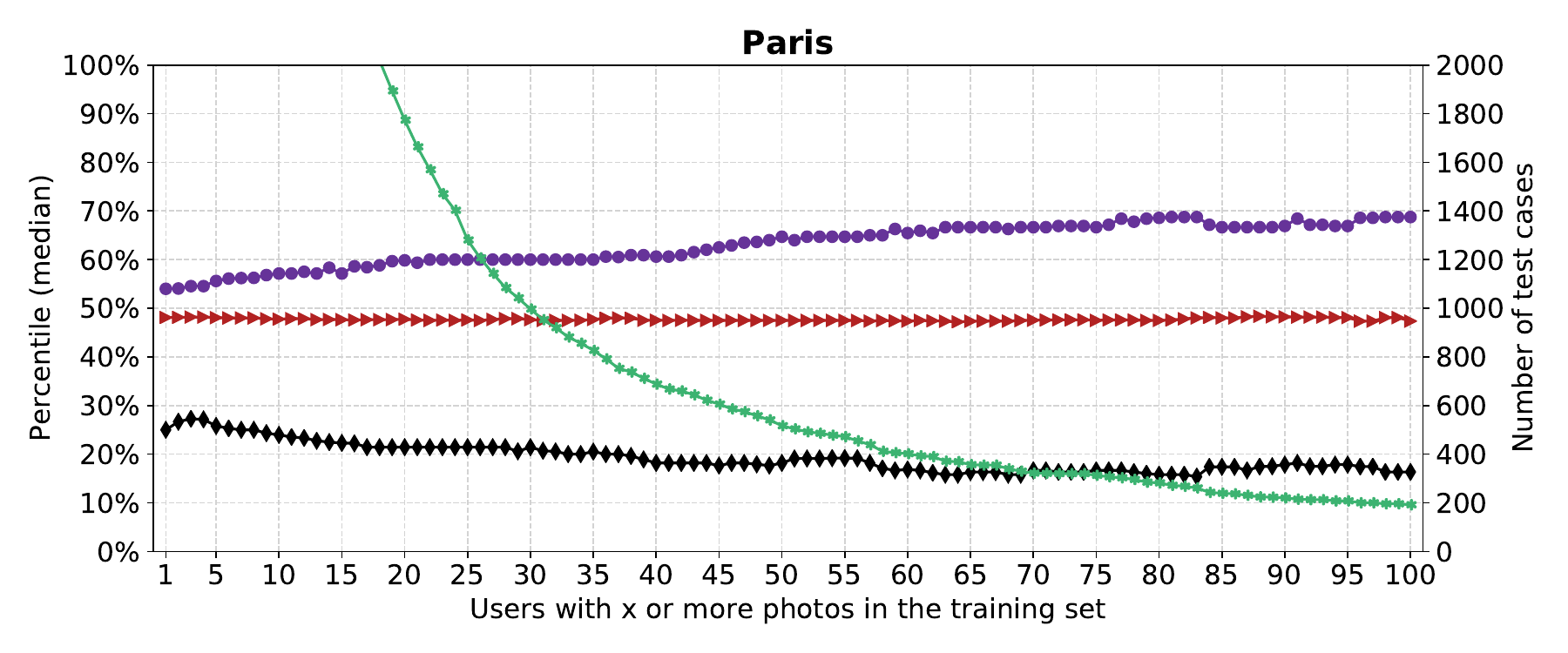}\\
    \includegraphics[width=\factor\linewidth]{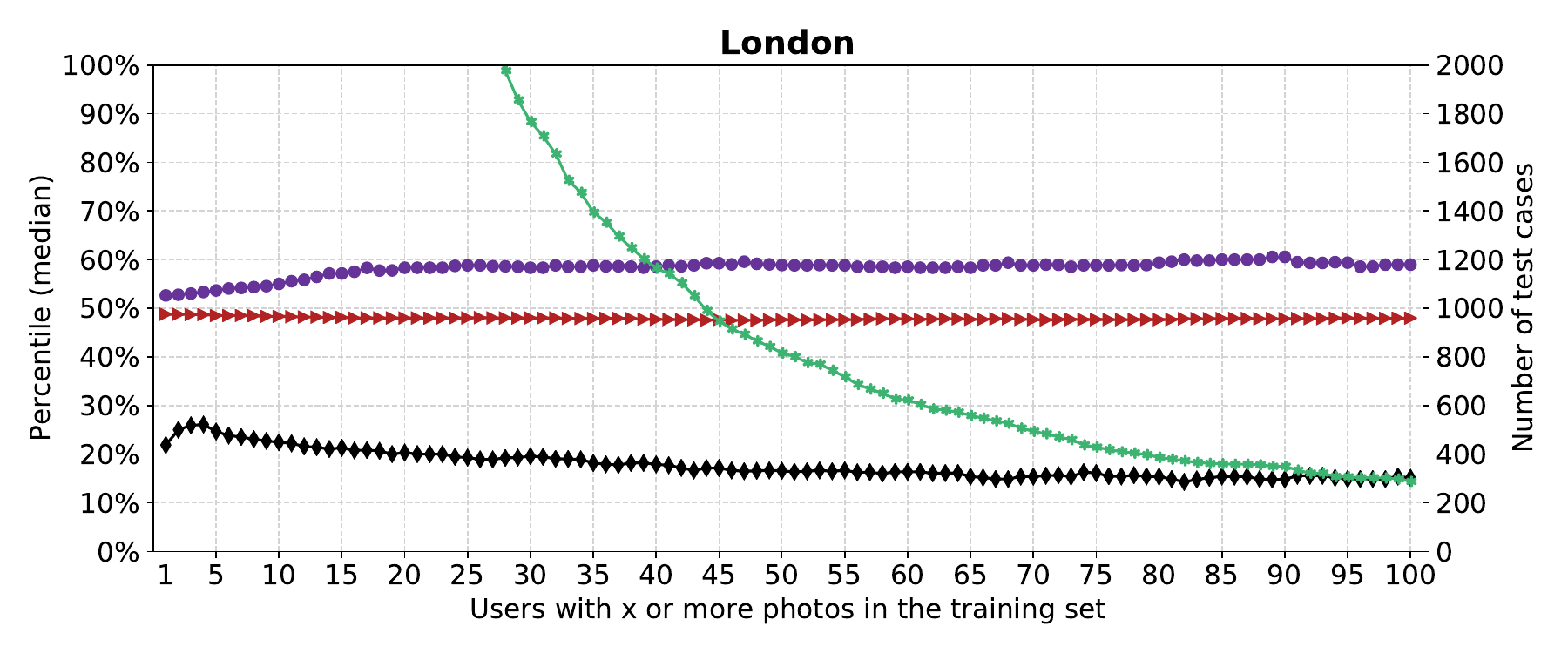}
    \caption{Average percentiles of the test photos, in \nyc, \paris and \london, for users with different number of photos in training. The green line represents the number of cases available below 2,000 (see the vertical axes in the right side)}
    \label{fig:results2}
\end{figure}

We also compared the performance of \oursystem and the two baselines regarding the amount of information available for training. For this purpose, we have tested the models with 100 different test sets for each city. Each test set was built as described in Section~\ref{sec:evaluation} but using only pairs $(\vec{u}, \vec{f})$ where the user has $x$ or more photos in the training set, with $x=\{1,2,\ldots,100\}$. Figures~\ref{fig:results} and~\ref{fig:results2} present the median percentile values of the test photo for each test set, where the $X$ axes represent the threshold used to filter out users regarding their amount of photos in the training set.


The variation in the amount of training information is barely relevant for the baselines. \emph{Random} keeps stable around the 50\%, while the \emph{centroid} is even worse, being always above that value. \oursystem exhibits by far the best performance. In fact, the performance of \oursystem increases when the model is applied to users of whom we have more information, as expected. This is reflected in the graphics, where the percentile score decreases (the lower the percentile, the better performance) as the test sets are built up of users with more photos (i.e., higher values in the $X$ axis).

However, it is important to consider the number of cases available in order to test the performance of the baselines and our model. Obviously, the more photos we require to include a user in a test set, the smaller (fewer users) it is. For this reason, the graphics in Figures~\ref{fig:results} and~\ref{fig:results2} also include the number of test cases (secondary $Y$ axes and green lines).

Notice that when the test sets include users with few photos (left side of the graphics) the median of the percentile increases for \oursystem, but it is still much better than the baselines. This indicates that the performance is good enough even for users in a situation near to a cold-start. In the next section we propose how to use our system for users in strictly cold-start situations, where we have no information at all.

The small number of cases in the test sets also explains the strange behavior of the \emph{centroid} and \oursystem in the city of \gijon: there is only a significant amount of test cases when including users with less than 25 photos. Consequently, their behavior in the \gijon dataset is not representative from that point on.

\subsection{The tastes of all users of a city about a restaurant}
\label{sec:el_perro_que_fuma}

\begin{figure}[tb]
    \centering
    \includegraphics[width=\textwidth]{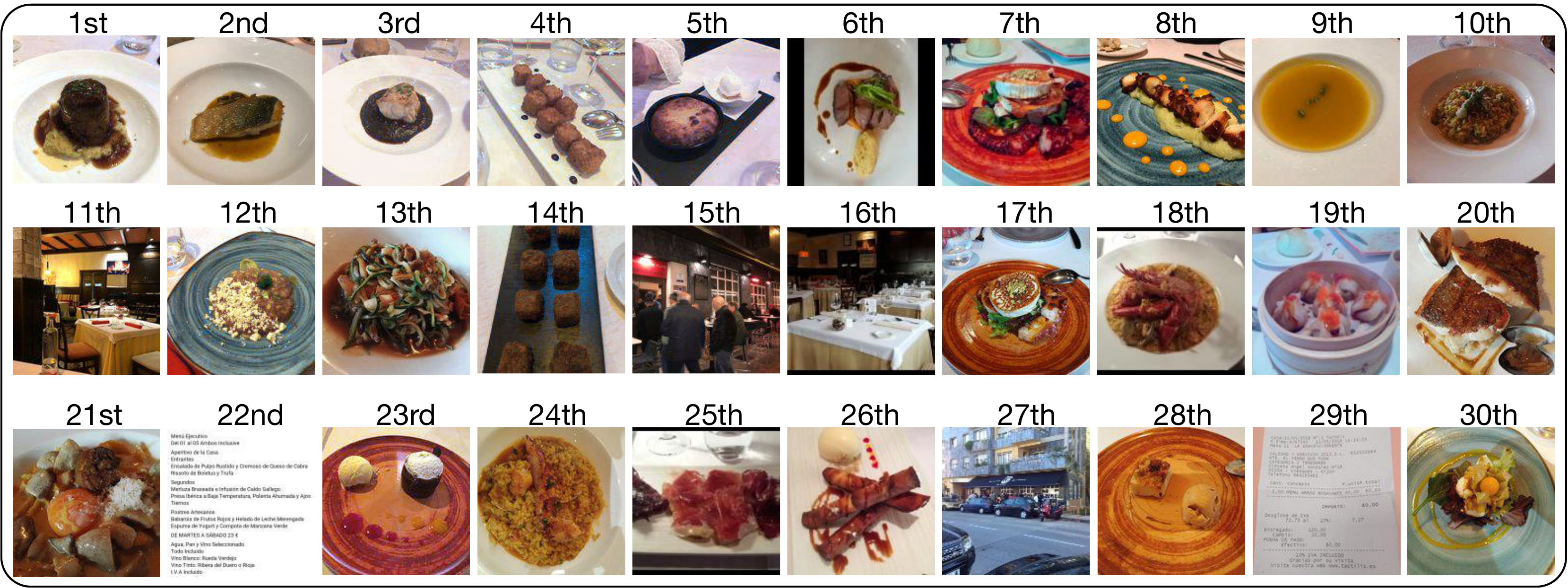}\vspace*{1em}
    \caption{The set of photos of one of the favorite restaurants of the group of authors of this paper. They are ordered using equation~(\ref{eq:SU_photo}), according to the preferences of the 5,139 users of TripAdvisor in \gijon . See the whole explanation in Section~\ref{sec:el_perro_que_fuma}. Some photos were slightly retouched in size and/or luminosity for a correct visualization in this paper}
    \label{fig:el_perro_que_fuma}
\end{figure}

Figure~\ref{fig:el_perro_que_fuma} shows the 30 photos  of ``El Perro que Fuma'' (one of the favorite restaurants of the group of authors of this paper in \gijon) available at the time of downloading the data. They are ordered by equation (\ref{eq:SU_photo}) using all the users of the \gijon dataset.

If we restrict the set of users to only those who took pictures in the restaurant, the results are similar.
If we look at the top-3 of the ranking, these are the variations: the first place would be for the photo that was in the ninth position before, the second would be the same, and the preferred photo for all clients (the one that occupied the first place) descends only to the third place.

Something sensibly different happens when we use the set of clients who visited the restaurant but declare that they do not like it. The favorite photo for them is the one that occupies the position 27 of 30. Curiously, this photo depicts the exterior area of the restaurant, taken from the opposite sidewalk and where you can barely see its facade, being hidden behind parked cars and a group of customers from the terrace of the restaurant.

We could use this approach to explain a recommendation for a new user with no historic information at all (cold-start). Provided we have a recommendation by any means to the new user, we could explain it using the top ranked photo(s) for the group of known users whose taste regarding that restaurant matches the recommendation (either positive or negative) made for the new user.

\section{Conclusions} \label{sec:conclusions}

In this paper we  present a method to explain the recommendations to the users of an RS; in particular, those in which the users share not only their tastes, but also their photos of items. We built a learning system (\oursystem) capable of predicting the photo of an item that the user would take in case of an eventual interaction with that item. In other words, the photo that reflects the most appealing aspect for the user. Users' photos serve to highlight those aspects.


The reason for pointing out users' photos as an element of explanation is that users disseminate photos to support their opinions in a way that seems unappealable. Thus, we believe that nothing can convince other users more than an image that could have been taken by themselves.

This method to convince users requires to learn a binary classification task aimed at finding out the authorship of photos. Formally, it works as an image recommender, but it has a peculiarity that marks the essence of the problem addressed in this paper. First of all we deal with photos (some of low quality) taken by users. Secondly, the photos are taken from a specific item. In other words, the photos have two dimensions that must be taken into account: their authorship and the item they portray.

To illustrate the performance of \oursystem in a real-world scenario, we used data taken from the TripAdvisor platform of six different cities. The datasets and the implementation are going to be publicly available with the publication of this paper, as was announced above. The results, compared with a pair of baseline methods, are excellent.

On the other hand, a comment on the evaluation of this system could be made. As with all RS, it could be argued that a fair evaluation should probably be established with a field test, asking users to assess the photos proposed by the model. The point is that the photos suggested by the model can be accepted by users as representative of their preferences, even better than those photos really taken by the user. As future work, we plan to consider how to apply synonymy of images to the context of this paper.

Last but not least, there is a side effect of \oursystem that we would like to highlight. We have seen that it allows to determine the most representative photos for a group of users within a set of photos. For example, the available photos of a restaurant can be \emph{democratically} ordered according to the preferences of the users. This option allows, for example, the owner of a restaurant to see the aspects of her business that stand out the most, not only the customers of the premises, but also all the possible customers. We believe that this is an important tool and, thus, its development is also part of our future work.

\section*{Acknowledgments}

This work was funded under grants TIN2015-65069-C2-2-R from the Spanish Ministry of the Economy and Competitiveness, and IDI-2018-000176 from the Principado de Asturias Regional Government, partially supported with ERDF funds.

We are grateful to NVIDIA Corporation for the donation of the Titan Xp GPU used in this research.

\bibliography{bibliography}

\end{document}